\documentclass[10pt,twocolumn,letterpaper]{article}
\usepackage[left=0.8125in,right=0.8125in,top=1in,textheight=8.875in,footskip=0.4in]{geometry}
\usepackage[T1]{fontenc}
\usepackage{times}
\usepackage{graphicx}
\usepackage{amsmath,amssymb,mathtools,bm}
\usepackage{booktabs,multirow,makecell,array,tabularx}
\usepackage[table,dvipsnames]{xcolor}
\usepackage{caption,subcaption}
\usepackage{enumitem,float,wrapfig,adjustbox}
\usepackage{algorithm,algorithmic}
\usepackage{microtype,xspace,etoolbox}
\usepackage[numbers]{natbib}
\definecolor{linkblue}{rgb}{0.21,0.49,0.74}
\usepackage[pagebackref,breaklinks,colorlinks,allcolors=linkblue]{hyperref}
\usepackage[capitalize,noabbrev]{cleveref}

\title{SuperFine3D: Efficient Global-Local Recursive Refinement for\\Extremely Detailed 3D Geometry Generation}
\author{Lihan Li\qquad Junwei Zhou\qquad Yu-wing Tai\textsuperscript{*}\\
Dartmouth College}
\date{}
\newcommand{\maketitlesupplementary}{\section*{Supplementary Material}}

\renewcommand*{\backref}[1]{}

\renewcommand*{\backrefalt}[4]{\ifcase#1\or #2\else #2\fi}
\makeatletter
\renewcommand{\@seccntformat}[1]{\csname the#1\endcsname.\hspace{0.5em}}
\renewcommand{\section}{\@startsection{section}{1}{\z@}{-10pt plus -2pt minus -2pt}{6pt plus 2pt minus 2pt}{\normalfont\large\bfseries}}
\renewcommand{\subsection}{\@startsection{subsection}{2}{\z@}{-8pt plus -2pt minus -2pt}{3pt plus 2pt minus 2pt}{\normalfont\normalsize\bfseries}}
\renewcommand{\subsubsection}{\@startsection{subsubsection}{3}{\z@}{-6pt plus -2pt minus -2pt}{3pt plus 2pt minus 2pt}{\normalfont\normalsize\bfseries}}
\renewcommand{\paragraph}{\@startsection{paragraph}{4}{\z@}{3pt plus 1pt minus 1pt}{-1em}{\normalfont\normalsize\bfseries}}
\renewcommand{\@maketitle}{\newpage\null\vskip .375in
\begin{center}
{\Large\bfseries\@title\par}\vskip .25in
{\large\begin{tabular}[t]{c}\@author\end{tabular}\par}
\vskip 4pt {\normalsize\url{https://lihhan.github.io/SuperFine3D-page/}\par}
\vskip .5em
\end{center}}
\makeatother
\renewenvironment{abstract}{\centerline{\large\bfseries Abstract}\vspace{12pt}\itshape\noindent}{\par\normalfont}

\begin{document}

\twocolumn[{%
\renewcommand\twocolumn[1][]{#1}%
\maketitle
\begin{center}
    \centering
    \includegraphics[width=\textwidth]{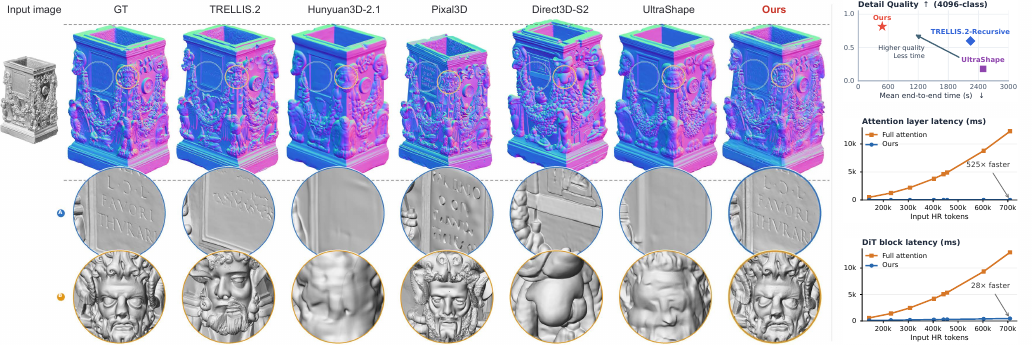}\\
    \captionsetup{type=figure}
    \caption{\textbf{High-resolution geometry with fine detail and efficient inference.}
    Left: input image. Middle: surface normals of GT, TRELLIS.2~\cite{xiang2025trellis2}, Hunyuan3D-2.1~\cite{hunyuan3d2025hunyuan3d21}, Pixal3D~\cite{li2026pixal3d}, Direct3D-S2~\cite{wu2025direct3ds2}, UltraShape~\cite{jia2025ultrashape}, and ours, with zooms A and B; only ours recovers the inscription and relief.
    Right: detail quality vs.\ end-to-end time for the only three methods that reach 4096-class resolution, and latency of one attention layer and one DiT block vs.\ high-resolution tokens; at 708k tokens ours is $525\times$ and $28\times$ faster than the full self-attention version of TRELLIS.2, Hunyuan3D-2.1 and the UltraShape refiner, run with FlashAttention-2~\cite{dao2024flashattention2}.}
    \label{fig:teaser}
\end{center}
}]
\begingroup\renewcommand{\thefootnote}{*}\footnotetext{Corresponding author.}\endgroup
\begin{abstract}
High-resolution images contain rich visual details that require
fine geometric structures for faithful 3D reconstruction. However, generating extremely detailed 3D geometry from images requires
substantial computational resources and training data, making
high-resolution geometry generation challenging for existing
state-of-the-art models. In this paper, we present SuperFine3D,
an efficient framework for generating extremely detailed 3D
geometry through global-local recursive refinement. SuperFine3D
learns a shared refinement operator from paired coarse-to-fine
representations and recursively reuses it to progressively
synthesize finer geometry beyond the training resolutions,
without requiring additional paired supervision at target
resolutions. To make recursive refinement computationally
efficient, we introduce spatial-cell attention, which restricts
high-resolution interactions to local spatial neighborhoods
while maintaining bounded communication across neighboring
cells. We further propose spatially anchored cross-scale
guidance, which leverages globally contextualized coarse geometry
to guide local detail synthesis and preserve structural
consistency. In addition, geometry-aware training objectives
encourage the refinement process to preserve stable geometric
structures while introducing fine-grained surface details.
Experiments demonstrate that SuperFine3D generates geometry
at a $4096^3$ decoder-grid resolution, corresponding to a sparse
$256^3$ latent lattice, using a shared refinement operator
trained only on $512^3$--$1024^3$ representations. Compared
with a full-attention recursive baseline, SuperFine3D achieves
a $9.4\times$ speedup in end-to-end inference
and reduces peak GPU memory usage by $23.0\%$, while improving
cross-scale structural consistency and producing cleaner
surfaces with finer geometric details.
\end{abstract}
\section{Introduction}
\label{sec:introduction}

Generating detailed 3D assets from images is an important capability
for digital content creation, simulation, and interactive
applications. Diffusion Transformers provide a scalable generative
architecture~\cite{peebles2023dit}, while structured latent
representations support detailed 3D generation~\cite{xiang2025structured3dlatentsscalable,xiang2025trellis2}. However, generating extremely detailed geometry at
ultra-high resolutions remains challenging. The increasing number
of active tokens makes high-resolution denoising computationally
expensive, while learning to generate finer geometry often requires
paired training representations at the target resolutions or
separately adapted generators. These requirements introduce
substantial computational and data demands.

In this paper, we address this problem by formulating high-resolution
3D generation as recursive geometric refinement. Instead of training
a separate generator for each target resolution, we learn a shared
refinement operator from paired coarse-to-fine representations.
The operator learns the geometric corrections associated with a
resolution-doubling transition and is recursively reused to
progressively synthesize finer geometry. This enables generation
beyond the training resolutions without requiring additional paired
supervision at each target resolution. As illustrated in
Fig.~\ref{fig:teaser}, SuperFine3D
generates extremely detailed 3D geometry that captures fine
structures and details from high-resolution image inputs.

The key insight of SuperFine3D is to combine computationally
efficient local refinement with globally informed structural
guidance. To achieve this, we introduce \textbf{spatial-cell
attention}, which restricts high-resolution interactions to tokens
within each spatial cell and a bounded neighborhood of adjacent
cells. By prioritizing communication across cell boundaries, it
preserves local geometric continuity while avoiding global
all-pairs attention over the entire high-resolution token set.
Complementing this local computation, we introduce
\textbf{spatially anchored cross-scale guidance}, which leverages
globally contextualized coarse geometry to guide local detail
synthesis. Spatially indexed cross-scale attention and continuous
interpolation anchors establish correspondence between irregular
sparse supports at different resolutions, allowing high-resolution
tokens to access both object-level context and spatially relevant
geometric information. Furthermore, we introduce geometry-aware
training objectives that distinguish stable geometric structures
from detail-bearing corrections, encouraging the refinement
operator to preserve existing geometry while introducing
fine-grained surface details. At inference time, each generated
latent provides both the structural guidance and sparse support
for the next refinement stage, forming a self-contained recursive
generation pipeline.

We instantiate SuperFine3D on the TRELLIS.2 sparse latent
representation and its pretrained geometry decoder. The shared
refinement operator is trained using paired representations at
$512^3$ and $1024^3$ geometry resolutions, and is recursively
applied to generate geometry at $2048^3$ and $4096^3$ resolutions.
The resulting $4096^3$ decoder-grid geometry corresponds to a
sparse $256^3$ latent lattice. Experiments demonstrate that
SuperFine3D achieves a $9.4\times$ speedup in end-to-end
inference and reduces peak GPU memory usage
by $23.0\%$ compared with a full-attention recursive baseline.
It also improves cross-scale structural consistency and produces
cleaner surfaces with finer geometric details.

Our main contributions are summarized as follows:
\begin{itemize}
    \item We present \textbf{SuperFine3D}, an efficient framework
    for ultra-high-resolution 3D geometry generation through
    recursive refinement. A shared refinement operator learned
    from $512^3$--$1024^3$ representations progressively synthesizes
    geometry up to $4096^3$ and beyond, without requiring additional paired
    supervision at the target resolutions.

    \item We introduce \textbf{spatial-cell attention} with bounded,
    boundary-prioritized cross-cell communication to efficiently
    denoise large high-resolution sparse supports.

    \item We propose \textbf{spatially anchored cross-scale guidance}
    and geometry-aware training objectives to ground local detail
    synthesis in coarse geometric structure, preserving structural
    consistency while introducing fine-grained details.
\end{itemize}
\section{Related Work}
\label{sec:related_work}

\begin{figure*}[t]
    \centering
    \scalebox{1.0}[1.0]{%
        \includegraphics[width=\linewidth]{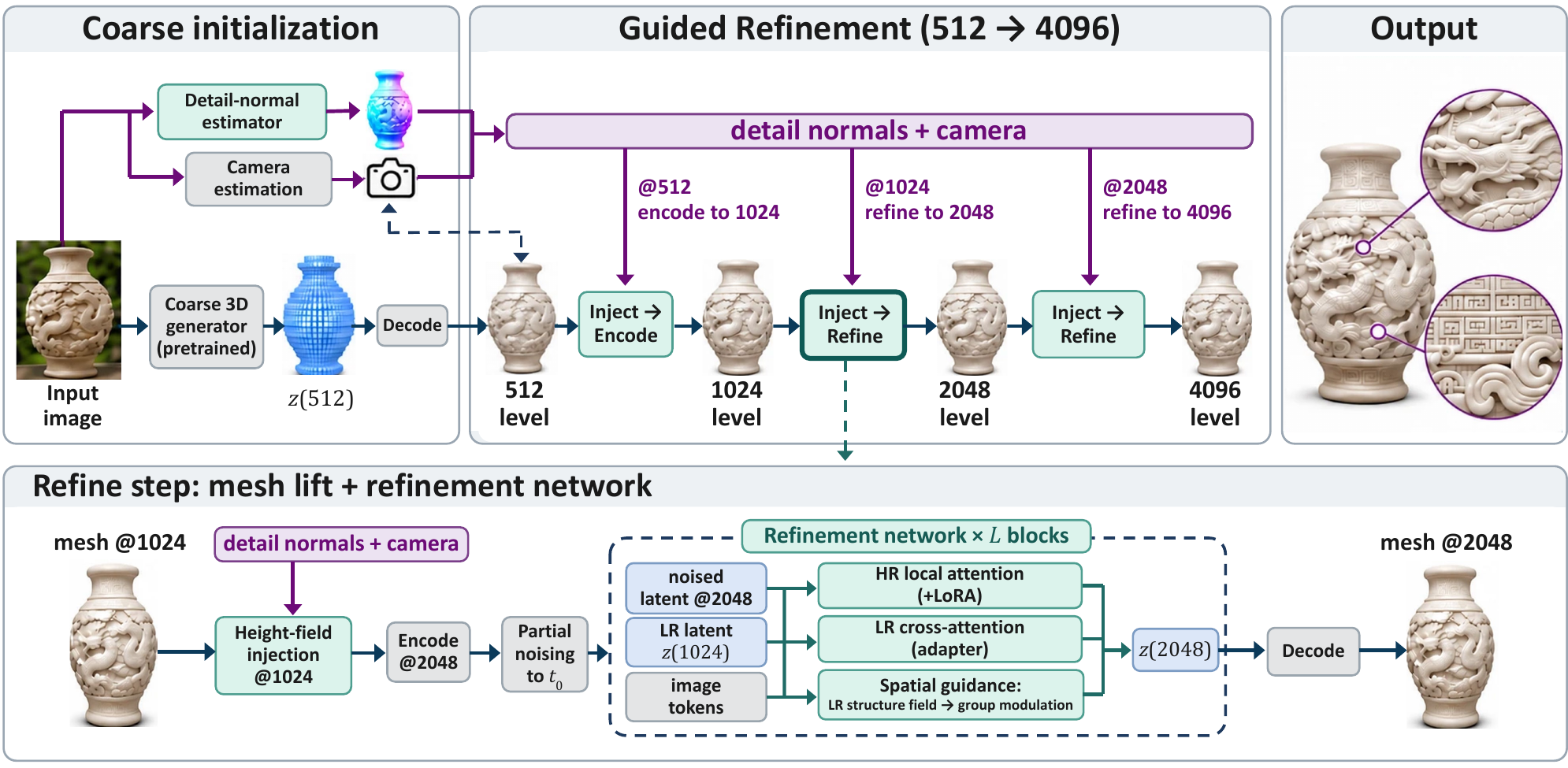}}
    \caption{
        Overview of SuperFine3D. A pretrained generator produces
        a coarse $512^3$ shape, from which the input camera is
        estimated. At every level, detail normals estimated from
        the input image are injected into the current mesh; the
        corrected mesh is encoded at twice the resolution,
        partially noised, and refined by the shared refinement
        network conditioned on the preceding latent and the
        image (bottom).
    }
    \label{fig:framework}
\end{figure*}

\subsection{Image-to-3D Generation and Representations}

Image-to-3D generation has moved from per-instance optimization to feed-forward reconstruction and native 3D generation.
Score distillation lifts 2D diffusion priors to 3D from
text~\cite{poole2022dreamfusion, lin2023magic3d} or a single
image~\cite{liu2023zero1to3, qian2024magic123,sam3d2025}; multi-view diffusion
improves cross-view consistency~\cite{shi2023MVDream, liu2023syncdreamer},
and large reconstruction models predict 3D directly from
images~\cite{hong2023lrm, tochilkin2024triposr}.
Representations range from triplanes~\cite{hong2023lrm, wang2024crm}
and 3D Gaussians~\cite{tang2024lgm, charatan2024pixelsplat} to
implicit functions~\cite{jun2023shap} and vector-set
latents~\cite{zhang20233dshape2vecset}, as used by
Hunyuan3D-2.1~\cite{hunyuan3d2025hunyuan3d21}; the attainable
geometric detail depends on both the representation and the
generator.

Sparse structured latents concentrate computation on active
regions \cite{xiang2025structured3dlatentsscalable, xiang2025trellis2},
and TRELLIS.2 pairs them with native geometry decoding
\cite{xiang2025trellis2}.
Scaling to $4096^3$, however, requires the generator itself to
synthesize features over far larger supports while keeping object
structure coherent.
Separately, monocular normal estimators recover sharp surface
detail from a single image
\cite{ye2024stablenormal, ye2025hi3dgen}, and Hi3DGen uses normals
as an intermediate representation for generation
\cite{ye2025hi3dgen}.
SuperFine3D combines the two: estimated normals constrain the
geometry at every level, and a shared refinement operator lifts
the result across resolutions.

\subsection{Coarse-to-Fine and Recursive Generation}
Coarse-to-fine generation separates object-level structure
from finer-scale synthesis.
Cascaded image diffusion models use successive super-resolution
stages to increase output resolution
\cite{ho2022cascaded, saharia2022photorealistic}.
Low-resolution structural conditioning can also guide localized
high-resolution image synthesis
\cite{zhang2025ultraimagegen}.
In 3D, Point-E generates and upsamples point
clouds~\cite{nichol2022point}, while LION models global and
pointwise latents hierarchically~\cite{zeng2022lion}.
Staged optimization offers another route to progressive detail
formation~\cite{lin2023magic3d,wang2023prolificdreamer}.
TRELLIS.2 uses a cascaded generation pipeline with finer sparse
supports~\cite{xiang2025trellis2}. UltraShape~\cite{jia2025ultrashape} instead refines a coarse shape once with
voxel-conditioned diffusion and extracts its surface at up to
$4096^3$.

UltraShape, however, decodes the whole object from a fixed budget
of 32{,}768 latent tokens on a $128^3$ query grid, so its
$4096^3$ grid only samples the implicit field more densely and
adds no generated detail.
Raising the latent itself to the $4096^3$ level would require
roughly an order of magnitude more tokens on a finer query grid,
which its full-attention denoiser cannot afford.
Recursive generation instead regenerates the shape at every
resolution, which adds a distinct requirement: the refinement rule
must remain valid as both its resolution and its conditioning
geometry change across stages.
SuperFine3D learns an explicitly LR-conditioned operator from
$512^3$--$1024^3$ pairs, reuses it at $2048^3$ and $4096^3$ with
local attention that keeps such token counts tractable, and
re-anchors every stage to the input image through explicit detail
injection, without paired supervision at the target resolutions.

\subsection{Efficient Attention in Diffusion Transformers}
Global self-attention in Diffusion Transformers
\cite{peebles2023dit} scales quadratically with token count.
FlashAttention \cite{dao2022flashattention} and RingAttention
\cite{liu2023ring} execute dense attention more efficiently without
changing its pairwise structure, while windowed attention
\cite{liu2021swin}, sparse convolutions, and octree attention
\cite{choy20194d, tang2020searching, wang2023octformer} exploit
spatial locality.
Direct3D-S2 uses spatial sparse attention for scalable 3D
generation~\cite{wu2025direct3ds2}.
On irregular 3D surface supports, fixed windows cover ever smaller
parts of the object as resolution grows; SuperFine3D therefore
pairs local spatial-cell attention with a globally contextualized
LR pathway.
\section{Method}
\label{sec:method}

\subsection{Image-Guided Recursive Refinement}
\label{sec:overview}

SuperFine3D learns a resolution-doubling refinement operator
from paired $512^3$--$1024^3$ representations and reuses the
same weights at higher resolutions (Fig.~\ref{fig:framework}).
We build on the frozen TRELLIS.2 generator, image encoder,
and geometry autoencoder~\cite{xiang2025trellis2}, where a shape at
decoder-grid resolution $R^3$ is a sparse latent
$z^{(R)}=(\mathcal S^{(R)},\mathbf F^{(R)})$
on a lattice of edge length $R/16$.

Inference starts from a $512^3$ shape (Pixal3D~\cite{li2026pixal3d}),
whose decoded mesh fixes the camera $\pi$, fitted once to the
silhouette and edges of the input image $I$.
Each level $R$ then alternates an image step, which injects
detail estimated from $I$ into the decoded mesh
(Sec.~\ref{sec:injection}), and a generative step, which
re-encodes the corrected mesh at twice the resolution and
refines it from an intermediate noise level:
\begin{align}
    x^{(2R)}
    &= \mathcal E_{2R}\bigl(\mathcal H_R(\mathcal D(z^{(R)});I,\pi)\bigr),
    \label{eq:lift}\\
    z^{(2R)}
    &= \operatorname{Sample}_{\theta}
    \bigl((1-t_0)\,x^{(2R)}+\sigma_{t_0}\epsilon;\;I,z^{(R)}\bigr).
    \label{eq:refinement_sampling}
\end{align}
Here $\mathcal D$ and $\mathcal E_{2R}$ are the frozen decoder and encoder, $\mathcal H_R$ is the injection at level $R$, $\sigma_t=\sigma_{\min}+(1-\sigma_{\min})t$, and sampling runs only the steps with $t\le t_0$.
At $1024^3$, Eq.~\eqref{eq:lift} alone gives the latent (posterior-mean encoder, as for the training
pairs); refinement then runs at $2048^3$ and $4096^3$ with shared $\theta$,
and a last injection at $4096^3$ yields the output mesh.
Since $\mathcal S^{(2R)}$ is the support of $x^{(2R)}$, the token layout
follows the image-corrected surface, while $z^{(R)}$ supplies the coarse
structure. With $t_0\le0.15$ (Sec.~\ref{sec:exp_setup}), inside the range of the geometric objectives ($t\le 0.5$, Sec.~\ref{sec:training}), each refinement is a single network evaluation.

\vspace{2mm}
\noindent\textbf{Image-guided detail injection.}
\label{sec:injection}
Image tokens reach the refinement network only at the
resolution of the image encoder ($64\times64$ tokens), too
coarse to specify relief at $2048^3$--$4096^3$.
We therefore transfer fine detail from the image explicitly,
in normal space.

\vspace{2mm}
\noindent\textbf{Detail normals.}
A one-step diffusion normal estimator (NiRNE~\cite{ye2025hi3dgen}), fine-tuned to resolve pixel-scale detail, runs on tiles of the upsampled input; for textured photographs, a mesh-conditioned variant separates relief from albedo.
With the rendered mesh normals $\mathbf N_M$, the estimate $\widehat{\mathbf N}$, and a low-pass filter $L_\kappa$ of $\kappa$ pixels, the target keeps the generated shape at coarse scales and takes the image at fine scales (basic form; gated form in the supplementary material):
\begin{equation}
    \mathbf T=\operatorname{normalize}\bigl(
    L_\kappa(\mathbf N_M)
    +g\,(\widehat{\mathbf N}-L_\kappa(\widehat{\mathbf N}))\bigr).
    \label{eq:normal_target}
\end{equation}

\vspace{2mm}
\noindent\textbf{Height field.}
Each visible vertex $\mathbf v$ moves along the viewing
direction $\boldsymbol\omega$ by a bounded displacement read
from a height field $s$ on the image grid:
\begin{equation}
    \mathbf v'=\mathbf v+h_R\,\gamma
    \tanh\!\bigl(s(\pi(\mathbf v))/\gamma\bigr)\,\boldsymbol\omega,
    \label{eq:height_field}
\end{equation}
where $h_R$ is the voxel width at level $R$ and $\gamma$ the
maximum displacement in voxels.
The field minimizes
$\sum_q(1-\mathbf n_q^{\top}\mathbf T_q)
+\alpha\sum_q w_q\,\rho(\nabla^2 s_q)+\beta\lVert s\rVert^2$,
where $\mathbf n_q$ is the displaced surface normal at pixel $q$,
$\rho$ is a Huber penalty, and $w_q$ relaxes smoothing on image
edges; it is solved coarse-to-fine.
Bounded displacements add relief at each level's scale without altering the global layout, and the next stage turns the corrected surface into a consistent latent.

\subsection{Resolution-Scalable Refinement}
\label{sec:local_refinement}

Two components let an operator trained at $1024^3$ run at $4096^3$: linear-cost HR attention and global structural guidance.

\noindent\textbf{Spatial-cell attention.}
Global HR self-attention becomes costly as the sparse
support grows.
We instead partition the HR lattice into cells of edge
length $b$.
Each query attends to all active tokens $\mathcal C(i)$
in its own cell $c(i)$ and a shared set
$\mathcal N_B(c(i))$ of at most $B$ tokens selected from
the 26 adjacent cells:
\begin{equation}
    a_i^H =
    \operatorname{Attn}
    \left(
        q_i,\,
        K_{\mathcal C(i)\cup\mathcal N_B(c(i))},\,
        V_{\mathcal C(i)\cup\mathcal N_B(c(i))}
    \right).
    \label{eq:local_attention}
\end{equation}
The cross-cell budget prioritizes face-adjacent boundary
tokens, with remaining slots assigned to spatially
distributed neighbors.
This maintains communication across cell boundaries
without global all-pairs interactions.

For $N_H$ active tokens and cell occupancies $n_c$,
the HR attention interaction cost is
\begin{equation}
    \mathcal O\!\left(\sum_c n_c^2+N_HB\right)
    \leq
    \mathcal O\!\left(N_H(b^3+B)\right).
    \label{eq:local_complexity}
\end{equation}
It therefore grows linearly with $N_H$. With $b{=}16$ and $B{=}128$, a $4096^3$ support of $N_H{=}2.6{\times}10^5$ tokens needs $2.6{\times}10^8$ query--key pairs instead of $7.0{\times}10^{10}$ for global attention, $265\times$ fewer.
Kernel details are given in the supplementary material.

\begin{figure*}[t]
    \centering
    \scalebox{1}[1.0]{%
        \includegraphics[width=\linewidth]{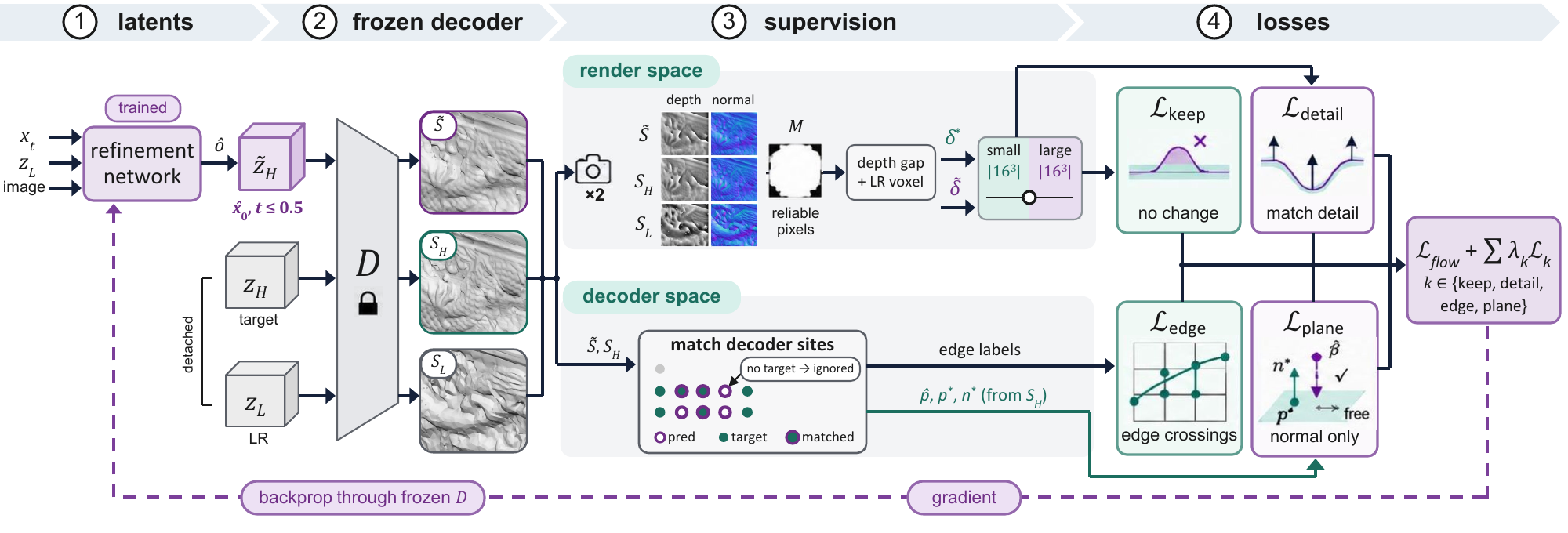}}
    \caption{\textbf{Training objective.}
    The prediction $\hat{z}_H$ ($=\hat{x}_0$, $t \le 0.5$), the target $z_H$ and the LR latent $z_L$ are decoded by a frozen decoder $D$.
    In render space, the depth corrections $\delta^{*}$ and $\hat{\delta}$ (in LR voxels, on reliable pixels $m$) drive $\mathcal{L}_{\text{keep}}$ (no change where the target is flat) and $\mathcal{L}_{\text{detail}}$ (match depth, slope and normal where it changes).
    In decoder space, matched sites give $\mathcal{L}_{\text{edge}}$ (balanced BCE on surface crossings) and $\mathcal{L}_{\text{plane}}$ (normal-only displacement).
    All terms are added to the flow-matching loss; gradients reach only the trainable adapters.}
    \label{fig:refinement_supervision}
\end{figure*}

\vspace{2mm}
\noindent\textbf{Spatially anchored cross-scale guidance.}
\label{sec:cross_scale}
Local HR attention alone provides limited object-level
context.
We therefore contextualize the preceding LR latent
globally and guide each HR token through spatially indexed
cross-attention and continuous interpolation anchors.
The former provides contextual information from the
corresponding coarse region, while the latter establishes
explicit correspondence between the two sparse lattices.

The LR stream incorporates image features through
cross-attention and then applies global self-attention.
Each HR cell reads all active contextualized LR tokens
in its corresponding coarse region $\mathcal P(c)$:
\begin{equation}
    a_i^L =
    \operatorname{Attn}
    \left(
        q_i,\,
        K^L_{\mathcal P(c(i))},\,
        V^L_{\mathcal P(c(i))}
    \right).
    \label{eq:lr_attention}
\end{equation}
Although the readout is spatially restricted, its LR
features carry context from the entire coarse shape.

To establish coordinate-level correspondence, we map
each HR token to the center-aligned LR lattice:
\begin{equation}
    \mathbf u_i
    =(\mathbf x_i^H+\tfrac12)\frac{r_L}{r_H}-\tfrac12,
    \label{eq:coordinate_mapping}
\end{equation}
where $r_L$ and $r_H$ are the latent-grid edge lengths.
We interpolate the active trilinear neighbors
$\mathcal V(i)$ to obtain an anchor and its coverage:
\begin{equation}
    \mathbf a_i^{\mathrm{anc}}
    =\sum_{j\in\mathcal V(i)}w_{ij}\phi(\mathbf f_j^L),
    \qquad
    c_i=\sum_{j\in\mathcal V(i)}w_{ij}.
    \label{eq:anchor}
\end{equation}
Here $\phi$ is a learned projection.
Missing neighbors contribute zero and weights are not
renormalized, so incomplete LR support attenuates the anchor.

As resolution increases, fixed-size HR cells cover
smaller object-space regions.
We additionally pool LR features and image context on
a fixed object-space grid and use the resulting descriptors
to modulate HR features.
This retains a consistent structural footprint across
recursive scales.

The readout and anchor are fused through a gated bottleneck,
with separate gates for the LR residual and the anchor, and
added to the HR attention update.

\subsection{Learning Structure-Preserving Refinement}
\label{sec:training}

We train on paired LR--HR latents (target LR latents and HR supports) with conditional flow
matching~\cite{lipman2023flow}, randomly perturbing the LR latent (local token
deletion and feature noise) so that the operator tolerates
generated LR inputs.
For $t\le 0.5$, we additionally decode the predicted clean
latent with the frozen decoder and supervise it geometrically,
separating target-supported detail from regions that should
remain stable (Fig.~\ref{fig:refinement_supervision}).

\paragraph{Preservation and detail supervision.}
Under shared views, let $d_L$, $d_H$, and $\widehat d_H$
denote depths of local surfaces decoded from the LR target,
HR target, and prediction.
We express cross-scale corrections in LR voxel units:
\begin{equation}
    \delta^\star=\frac{d_H-d_L}{h_L},
    \qquad
    \widehat\delta=\frac{\widehat d_H-d_L}{h_L},
    \label{eq:normalized_refinement}
\end{equation}
where $h_L$ is the LR voxel width.
A detached mask $m$ selects valid depth correspondences
using visibility, target-normal consistency, and bounded
cross-scale displacement.
Target corrections define complementary weights:
\begin{equation}
    w_{\mathrm{keep}}=m e^{-|\delta^\star|/\tau},
    \qquad
    w_{\mathrm{detail}}=m-w_{\mathrm{keep}}.
    \label{eq:refinement_weights}
\end{equation}
For the weighted mean
$\langle x\rangle_w=\sum wx/\max(\sum w,1)$,
preservation penalizes changes beyond the target-required
magnitude and tolerance $\eta$:
\begin{equation}
    \mathcal L_{\mathrm{keep}}
    =
    \left\langle
        [|\widehat\delta|-|\delta^\star|-\eta]_+^2
    \right\rangle_{w_{\mathrm{keep}}}.
    \label{eq:keep_loss}
\end{equation}
Detail supervision instead matches the direction and
spatial variation of the required correction:
\begin{equation}
    \mathcal L_{\mathrm{detail}}
    =
    \left\langle
        \rho_{\mathrm d}(\widehat\delta-\delta^\star)
    \right\rangle_{w_{\mathrm{detail}}}
    +\mathcal L_\nabla
    +\lambda_n\mathcal L_{\mathrm{normal}}.
    \label{eq:detail_loss}
\end{equation}
Here $\rho_{\mathrm d}$ is smooth-$L_1$,
$\mathcal L_\nabla$ matches correction gradients on valid
pixel pairs, and $\mathcal L_{\mathrm{normal}}$ penalizes
reliable surface-normal disagreement.

\paragraph{Decoder-space supervision.}
At matched decoder sites, a class-balanced loss
$\mathcal L_{\mathrm{edge}}$ supervises surface crossings, and a
tangent-plane loss $\mathcal L_{\mathrm{plane}}$ penalizes only
normal-direction vertex offsets.

The complete objective is
\begin{equation}
\begin{aligned}
    \mathcal L ={}&
    \mathcal L_{\mathrm{FM}}
    +\lambda_{\mathrm k}\mathcal L_{\mathrm{keep}}
    +\lambda_{\mathrm d}\mathcal L_{\mathrm{detail}}\\
    &+\lambda_{\mathrm e}\mathcal L_{\mathrm{edge}}
    +\lambda_{\mathrm p}\mathcal L_{\mathrm{plane}}.
\end{aligned}
\label{eq:total_loss}
\end{equation}
Only the HR attention LoRA adapters~\cite{hu2022lora} and the
LR conditioning modules are trained.

\section{Experiments}
\label{sec:experiments}
\subsection{Setup and Resolution Generalization}
\label{sec:exp_setup}
The refinement operator is initialized from the TRELLIS.2 $1024^3$ flow model and trained on 10,000 assets with paired $512^3$--$1024^3$ posterior-mean latents: HR attention LoRA (rank 16, all 30 blocks) is adapted first, then optimized jointly with the LR conditioning modules.
To resolve pixel-scale relief, the detail-normal estimator is fine-tuned from NiRNE~\cite{ye2025hi3dgen} on 1,961 close-up renders ($2$--$4\times$ camera zoom) of 480 training assets, disjoint from all test assets, upsampled as at inference and supervised by a cosine loss with a high-pass term.
All results start from the $512^3$ latent of Pixal3D~\cite{li2026pixal3d}, except the GT@512 row of Table~\ref{tab:quality_comparison}, which starts from the reference shape at $512^3$ to isolate SuperFine3D from the coarse generator.
Table~\ref{tab:quality_comparison} uses $t_0=0.15$/$0.05$ at $2048^3$/$4096^3$, $g=1$, and the gated target for all inputs; figure settings, including flow alignment and test-time averaging, are in the supplementary material.

\begin{figure}[t]
    \centering
    \scalebox{0.5}[0.5]{%
        \includegraphics[width=\textwidth]{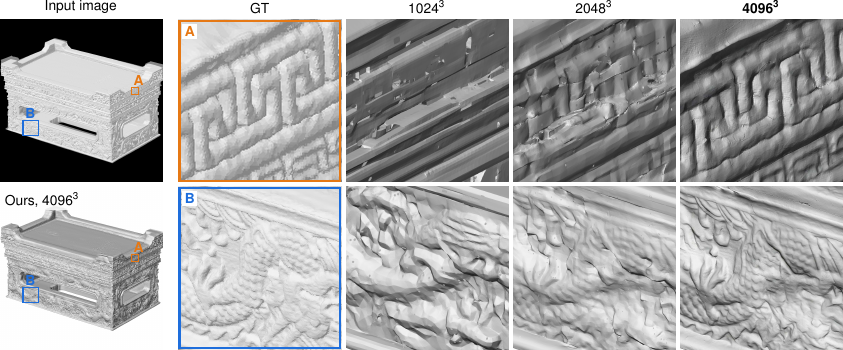}}\\
    \caption{
        Generation beyond training resolutions.
    }
    \label{fig:resolution_progression}
\end{figure}

\vspace{2mm}
\noindent\textbf{Beyond the training resolution.}
\label{sec:resolution_generalization}
With the same weights at both refinement stages, the fret pattern and scroll ornament sharpen at each level from $1024^3$ to $4096^3$ (Fig.~\ref{fig:resolution_progression}): one shared operator, not separately trained generators, carries detail beyond its training resolution.

\subsection{Comparison with Existing Methods}
\label{sec:method_comparison}

\begin{figure*}[t]
    \centering
    \scalebox{1.0}[1.0]{\includegraphics[width=\textwidth]{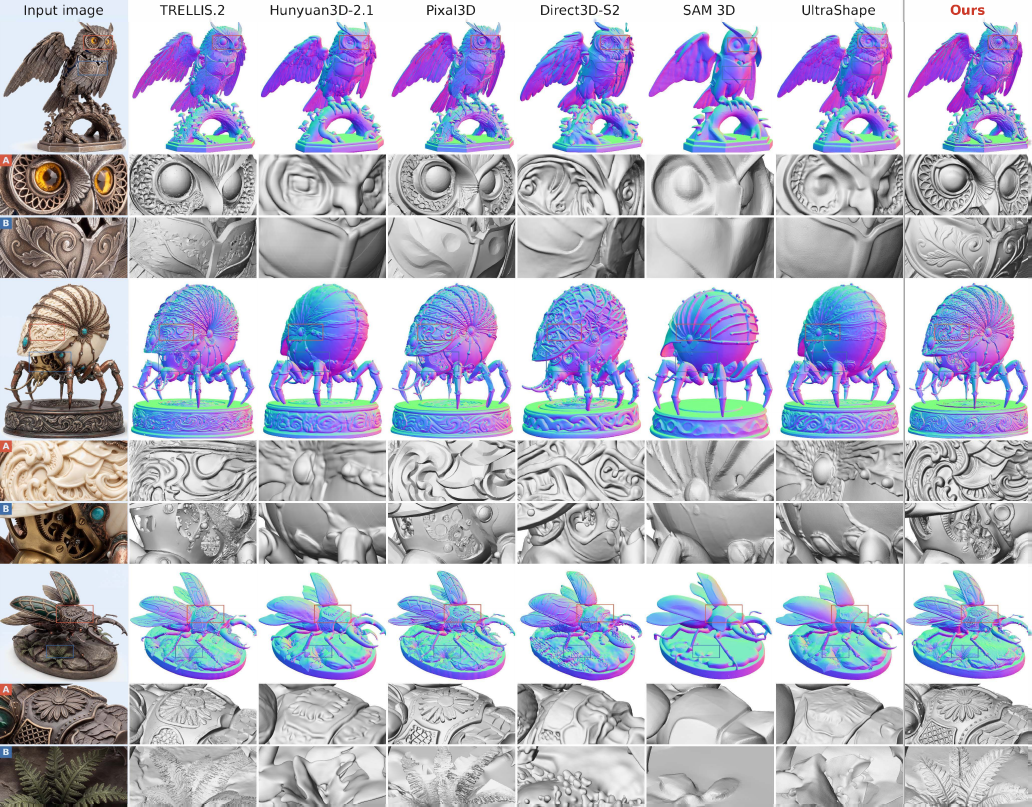}}
\caption{
    \textbf{Comparison of fine geometric details.}
    Inputs are $4096$\,px images generated with Nano Banana Pro~\cite{nanobananapro2025}, identical for all methods.
}
\vspace{-0.1in}    \label{fig:baseline_detail_comparison}
\end{figure*}

We compare against TRELLIS.2~\cite{xiang2025trellis2}, Hunyuan3D-2.1~\cite{hunyuan3d2025hunyuan3d21}, Pixal3D~\cite{li2026pixal3d}, Direct3D-S2~\cite{wu2025direct3ds2} ($1024^3$), and UltraShape~\cite{jia2025ultrashape}, each at its highest released setting.
All methods receive identical input images.

Table~\ref{tab:quality_comparison} reports averages over 200 common Toys4K~\cite{stojanov2021using} assets.
CD-L1 and F-score measure distance to and agreement with the reference surface;
Ang.\ and Corr compare normals with the reference in the input view (mean angle; correlation of 4--16 and 16--64\,px detail);
the reference-free Detail Quality (DQ)~\cite{meshybench2026} measures local detail regularity.
Full definitions, settings, and resolution fallbacks are in the supplementary material.

SuperFine3D attains the highest detail quality (DQ 0.647 at $4096^3$, $+18.9\%$ over TRELLIS.2) and follows the input image most closely: its input-view normal error is $27.5^\circ$ against $33.7^\circ$ for the best baseline, and its 4--16\,px detail correlation is about $10\times$ higher (0.368 vs.\ 0.038). Started from the GT $512^3$ shape, SuperFine3D$^\dagger$ reaches a CD-L1 of 0.657 and an $\mathrm{F}_{.002}$ of 93.2, so the remaining shape error stems from the coarse start.

\begin{table}[t]
    \centering
    \caption{
        \textbf{Geometric quality.}
        Distances are scaled by $10^3$; F-scores are percentages.
        Best non-oracle values are bold; \\
        $^\dagger$starts from the reference shape at $512^3$ (GT@512, oracle)
    }
    \label{tab:quality_comparison}
    \scriptsize
    \setlength{\tabcolsep}{2.5pt}
    \begin{tabular*}{\linewidth}{
        @{\extracolsep{\fill}}lccccccc@{}
    }
        \toprule
        Method & Res. &
        CD-L1 $\downarrow$ &
        Ang.($^\circ$)\ $\downarrow$ &
        $\mathrm{F}_{.002}$ $\uparrow$ &
        DQ $\uparrow$ &
        \multicolumn{2}{c@{}}{Corr.(px)\ $\uparrow$} \\
        \cmidrule(l){7-8}
        & & & & & & 4--16 & 16--64 \\
        \midrule
        TRELLIS.2 & 1536
        & 15.711 & 35.40 & 17.075 & 0.5444 & 0.002 & 0.105 \\
        Hunyuan3D-2.1 & 1024
        & 16.115 & 38.42 & 14.204 & 0.3680 & 0.006 & 0.110 \\
        Pixal3D & 1536
        & 17.761 & 33.70 & 14.197 & 0.5319 & 0.038 & 0.268 \\
        UltraShape & 4096
        & 17.381 & 35.76 & 12.083 & 0.1708 & 0.014 & 0.166 \\
        \midrule
        SuperFine3D & 2048
        & \textbf{15.639} & \textbf{27.44} & 17.107 & 0.5957
        & 0.364 & 0.501 \\
        SuperFine3D & 4096
        & 15.660 & 27.46 & \textbf{17.123} & \textbf{0.6474}
        & \textbf{0.368} & \textbf{0.518} \\
        \midrule
        SuperFine3D$^\dagger$ & 4096 & 
        0.657 & 13.77 & 93.179 & 0.6638 & 0.556 & 0.784 \\
        \bottomrule
    \end{tabular*}
\end{table}


\vspace{2mm}
\noindent\textbf{Fine geometric detail.}
Figure~\ref{fig:baseline_detail_comparison} compares the methods
on three assets with intricate surface patterns and fine geometric
structures, each a $4096$\,px image generated with Nano Banana
Pro~\cite{nanobananapro2025}.
TRELLIS.2 yields granular relief, while Hunyuan3D-2.1 and
SAM 3D oversmooth fine structures.
Pixal3D and Direct3D-S2 recover some local patterns but fail to
consistently preserve their geometric structures, and UltraShape
also loses many fine details.
In contrast, SuperFine3D recovers the openwork eye rims, leaf and
shell scrolls, gears, and fern fronds where the image shows them.
These results demonstrate the effectiveness of our framework
in recovering fine, image-specific geometry while preserving
the overall structure of complex objects.

\vspace{2mm}
\noindent\textbf{Comparison at matched output resolution.}
UltraShape and TRELLIS.2--Recursive are the only baselines that reach a $4096^3$-class grid.
TRELLIS.2--Recursive reuses the released TRELLIS.2 flow models~\cite{xiang2025trellis2} at progressively finer grids, with supports predicted by the preceding stage's decoder.
In Fig.~\ref{fig:teaser}, UltraShape at $4096^3$ smooths the inscription (A) and the mask (B) away, whereas SuperFine3D keeps the lettering legible and the face, curls, and beard in the ground-truth layout; Fig.~\ref{fig:teaser} (top right) shows that TRELLIS.2--Recursive also trails in detail quality at a higher cost.
This indicates that the gain comes from image-guided injection and LR-guided refinement rather than grid resolution alone.


\subsection{Computational Efficiency and Scalability}
\label{sec:efficiency}

We benchmark the same asset on an NVIDIA RTX PRO 6000 for the three methods that reach a $4096^3$-class grid, each with its own default sampling settings.
Each run starts on an idle GPU and includes image conditioning, coarse generation, all refinement stages, geometry decoding, and mesh export; for SuperFine3D this also includes camera fitting and detail injection.
UltraShape extracts its surface with a memory-bounded, chunked decoder.\footnote{UltraShape's official decoder evaluates a dense implicit field over the full extraction grid on the GPU. This already requests 413\,GiB at $2048^3$ and would need about 3\,TB at $4032^3$, so we keep the field in host memory and evaluate it in GPU-sized chunks; the extracted field is identical to the official one.}
Table~\ref{tab:efficiency} reports latency and peak allocated GPU memory.

\begin{table}[t]
    \centering
    \caption{\textbf{End-to-end inference at $4096^3$-class resolution.}
    Time and peak allocated GPU memory on the same input.}
    \label{tab:efficiency}
    \small
    \setlength{\tabcolsep}{3pt}
    \begin{tabular*}{\linewidth}{@{\extracolsep{\fill}}lrr@{}}
        \toprule
        Method & Time (s) $\downarrow$ & Memory (GiB) $\downarrow$ \\
        \midrule
        TRELLIS.2--Rec.          & 1654.14 & 63.07 \\
        UltraShape     & 2214.71 & -- \\
        Ours                     & 176.19 & 48.54 \\
        \bottomrule
    \end{tabular*}
\end{table}

At $4096^3$, SuperFine3D is $9.4\times$ faster than TRELLIS.2--Rec.\ with 23.0\% less peak GPU memory, and about $12.6\times$ faster than UltraShape, whose extraction fits on the GPU only by keeping the field in host memory.

To compare denoising on identical workloads,
Figure~\ref{fig:teaser} measures the first DiT block
across eight assets spanning 143,440--708,592 HR tokens.
Each pair uses identical HR coordinates, noise, image conditioning,
batch size one, and BF16 activations on the same GPU model.
We report the median of nine synchronized measurements after two
warm-up runs. Ours includes HR--LR attention and LR feature transfers. Our block latency stays below 0.5\,s on all tested workloads and is $28\times$ lower than full attention at 708k tokens.

\begin{figure*}[t]
    \centering
    \scalebox{1}[1.0]{\includegraphics[width=\textwidth]{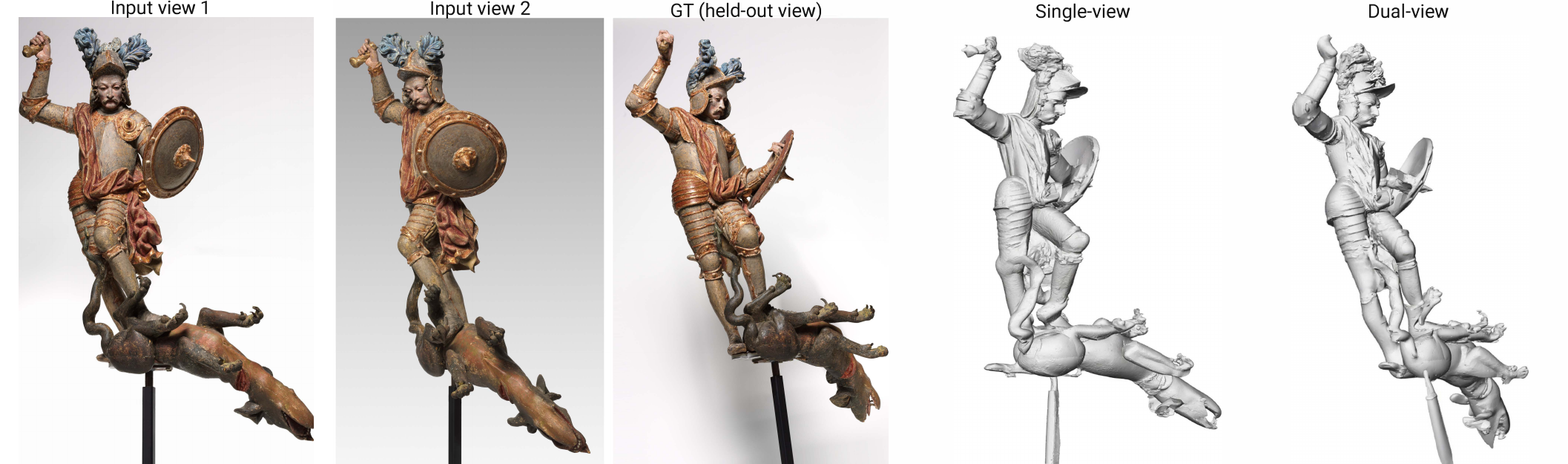}}\\
    \caption{
        \textbf{Novel-view comparison.} Inputs from the front-left and the left; GT is a held-out photo from the front-right, $62^\circ$ and $112^\circ$ from the inputs. With the same pipeline, a single view stretches the dragon sideways; two views recover its pose.
    }
    \label{fig:multi_input}
\end{figure*}

\subsection{Improving Structure with Multiple Views}
\label{sec:multi_input}

SuperFine3D also accepts multiple posed images.
We estimate cameras and the focal length with MASt3R~\cite{mast3r},
average pixel-aligned features across views in
the coarse generator, and refine the result to $4096^3$ with
the unchanged pipeline (detail injection uses the first view).
This targets errors that refinement cannot fix: a wrong coarse
layout propagates through every later stage, whereas a second
view constrains depth and occluded parts from the start.

In Fig.~\ref{fig:multi_input}, the inputs are $50^\circ$ apart and a held-out photo is the reference. With one view, the knight stands upright and the dragon is stretched sideways; with two, both match the held-out photo, raising silhouette IoU from 0.22 to 0.64 (0.48 to 0.81 on a second held-out photo).

\subsection{Ablation Studies}
\label{sec:ablations}

We investigate the roles of image-guided detail injection, the fine-tuned detail-normal estimator, and geometry-aware supervision in high-resolution recursive refinement.

\begin{figure}[t]
    \centering
    \includegraphics[width=\linewidth]{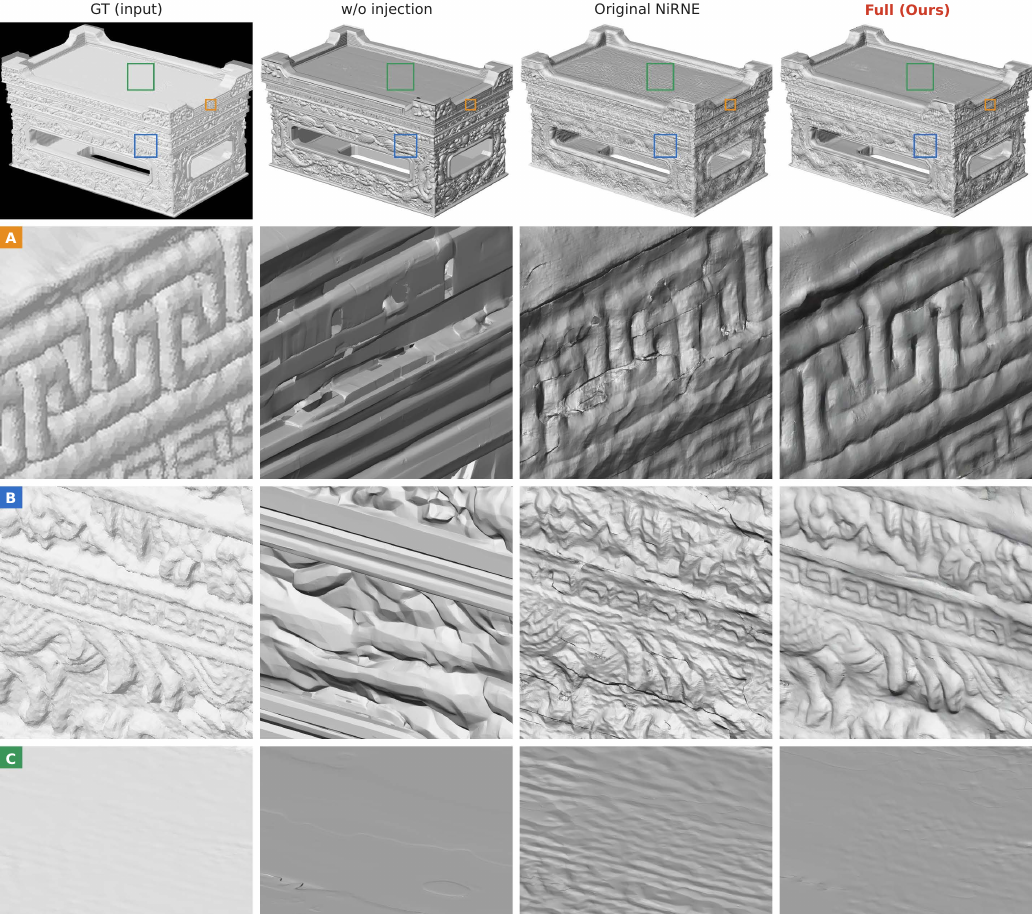}\\
    \caption{\textbf{Ablation of detail injection and the detail-normal estimator at $4096^3$.}
    Left to right: ground-truth render (input), w/o injection, original NiRNE, and the full model. Close-ups: (A) large fret pattern, (B) fine fret band, (C) flat lid.}
    \label{fig:injection_ablation}
\end{figure}

\vspace{2mm}
\noindent\textbf{Detail injection.}
On the asset of Fig.~\ref{fig:resolution_progression}, we remove detail injection at every level ($512^3$ to $4096^3$) while keeping the same initialization, refinement weights, and sampling schedule.
Without injection, the box, its panels, and the coarse ornament layout are preserved, but refinement alone cannot recover image-specific relief: the fret pattern (A) collapses into parallel bands and broken slabs, and the fret band and scroll (B) become faceted lumps (Fig.~\ref{fig:injection_ablation}).
With injection, both patterns follow the input in layout and depth.
Detail injection thus supplies the image-specific relief, while the shared refinement operator carries the shape across scales.

\begin{figure}[t]
    \centering
    \scalebox{1}[1.0]{%
        \includegraphics[width=\linewidth]{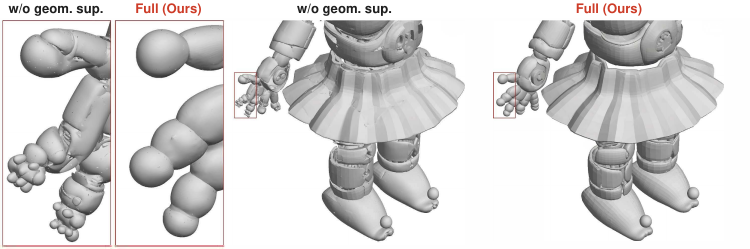}}\\
    \caption{\textbf{Effect of geometry-aware supervision.} Matched $4096^3$ hand close-ups show that it suppresses unnecessary surface changes while retaining distinct finger geometry.}
    \label{fig:ablation_details}
\end{figure}

\vspace{2mm}
\noindent\textbf{Detail-normal estimator.}
Replacing the fine-tuned estimator with the original NiRNE~\cite{ye2025hi3dgen} weights (Fig.~\ref{fig:injection_ablation}, third column) still recovers the large fret pattern (A), but blurs the fine fret band (B) and adds ripples to the flat lid (C). Fine-tuning thus resolves small, low-contrast ornament and suppresses spurious relief.

\vspace{2mm}
\noindent\textbf{Geometry-aware supervision.} Each level refines the previous output, so spurious structure persists across levels. Without the geometric objectives of Eq.~\eqref{eq:total_loss} (flow matching kept), fingertips at $4096^3$ branch into extra lumps; geometry-aware supervision suppresses them while keeping the fingers separated (Fig.~\ref{fig:ablation_details}). Its preservation term holds stable regions fixed, while the detail, surface-crossing, and tangent-plane terms match the target's corrections, surface crossings, and normal vertex offsets.

\vspace{2mm}
\noindent\textbf{Limitations.} The quality of recursive refinement depends on the geometric cues available from the input images and the initial coarse geometry. In particular, estimated surface normals may not fully capture fine geometric structures, while substantial deviations in the initial geometry can make it challenging to maintain global structural consistency. These challenges are more relevant to single-view inputs, where visual evidence is inherently limited. Exploring more reliable geometric cues and stronger structural guidance is a promising direction for future work.

\section{Conclusion}
\label{sec:conclusion}
We presented SuperFine3D, an efficient framework for generating extremely detailed 3D geometry from images through global-local recursive refinement. To enable efficient high-resolution refinement, we introduced spatial-cell attention for local computation and spatially anchored cross-scale guidance to preserve global structure while synthesizing local details. Geometry-aware training objectives further encourage structural preservation and fine-grained detail synthesis. Experiments demonstrate that SuperFine3D generates geometry at a $4096^3$ decoder-grid resolution, achieving a $9.4\times$ speedup and reducing peak GPU memory usage by $23.0\%$ compared with a full-attention recursive baseline. Future work will explore scaling SuperFine3D to larger and more complex 3D structures.


{\small
\bibliographystyle{ieeenat_fullname}
\bibliography{main}
}
\clearpage
\clearpage
\maketitlesupplementary
\renewcommand{\thesection}{\Alph{section}}
\setcounter{section}{0}
\renewcommand{\theequation}{S\arabic{equation}}
\setcounter{equation}{0}
\renewcommand{\thefigure}{S\arabic{figure}}
\setcounter{figure}{0}
\renewcommand{\thetable}{S\arabic{table}}
\setcounter{table}{0}

\section{Scope and Resolution Conventions}
\label{supp:scope}
This supplement details the image-guided recursive inference procedure and
its coarse start, the detail-normal estimator and height-field injection,
multi-view input, geometric supervision and training, baseline
configurations, evaluation metrics with additional comparisons, and the
efficiency analysis.
All resolutions refer to geometry unless otherwise stated. Figures and tables numbered without the prefix S are those of the main paper.
For the TRELLIS.2~\cite{xiang2025trellis2} representation, a geometry grid of edge length $R$ is
represented by a sparse latent lattice of edge length $r=R/16$.
Consequently, $512^3$, $1024^3$, $2048^3$, and $4096^3$ geometry correspond
to $32^3$, $64^3$, $128^3$, and $256^3$ latent lattices.
Only active lattice sites are instantiated, so computation follows the
object surface rather than the full $R^3$ volume.

The refinement operator is trained only on full-object $512^3$--$1024^3$
pairs and is deployed with shared weights at $2048^3$ and $4096^3$.
At every level, an explicit, training-free injection step transfers fine
detail from the input image, so each new resolution is grounded in the
image. Each level synthesizes new latent features on a finer lattice, which
adds geometry that a fixed continuous field sampled more densely cannot
contain.

\section{Image-Guided Recursive Inference}
\label{supp:inference}
\subsection{Pipeline}
Let $\mathcal D$ and $\mathcal E_R$ be the frozen decoder and the encoder at
geometry resolution $R$, and $\mathcal H_R$ the image-guided injection at
level $R$ (Sec.~\ref{supp:injection}). Inference proceeds as
\begin{equation}
\begin{aligned}
 z^{(1024)}&=\mathcal E_{1024}\big(\mathcal H_{512}(\mathcal D(z^{(512)}))\big),\\
 x^{(2R)}&=\mathcal E_{2R}\big(\mathcal H_R(\mathcal D(z^{(R)}))\big),\\
 z^{(2R)}&=\mathrm{Sample}_\theta\big(x^{(2R)};\,I,z^{(R)}\big),
 \quad R\in\{1024,2048\}.
\end{aligned}
\end{equation}
The coarse latent $z^{(512)}$ is the $512^3$ shape of Pixal3D~\cite{li2026pixal3d} in the TRELLIS.2 latent space (Sec.~\ref{supp:start}). The first lift to $1024^3$ is a deterministic encoding of the
image-corrected $512^3$ mesh with the posterior-mean encoder that also builds
the training pairs, so the first refinement stage starts from the same
representation it was trained on. Both subsequent lifts use the same
refinement weights $\theta$. For the second lift, the injected mesh is the
decoded $2048^3$ refinement result.

\paragraph{Partial noising.}
The encoded latent $x^{(2R)}$ is normalized as in training and noised to
$t_0$,
\begin{equation}
 x_{t_0}=(1-t_0)\,x^{(2R)}+\sigma_{t_0}\epsilon,\qquad
 \sigma_{t_0}=\sigma_{\min}+(1-\sigma_{\min})t_0 ,
\end{equation}
and the sampler continues the 12-step schedule (time rescaling factor 3)
from $t_0$. We use $t_0=0.15$, and $t_0=0.05$ at the $4096^3$ stage for the
results listed in Sec.~\ref{supp:settings}; both lie inside the range
$t\le0.5$ on which the geometric objectives are trained (Sec.~\ref{supp:losses}),
and from either the schedule reduces to a single Euler step to $t=0$, so each stage needs one
evaluation of the refinement network, because the encoded latent already
carries the image-corrected structure. The sparse support of the
stage is that of $x^{(2R)}$, so the token layout already follows the
image-corrected surface; the LR condition $z^{(R)}$ is the preceding latent
before injection. Refinement uses guidance scale $1$ (conditional prediction
only).

\paragraph{Encoding at $4096^3$.}
We encode at $4096^3$ with 64-bit coordinate codes and overlapping slabs
along one axis, which bounds encoder memory while covering the complete
surface.

\paragraph{Output.}
After the last stage, the $4096^3$ latent is decoded and a final injection at $4096^3$ is applied unless the resolution gate (Sec.~\ref{supp:gate}) skips it; small boundary loops (at most 60 edges and $8/4096$
extent) are capped and T-junctions closed. All reported detail is therefore
produced by generation and injection.

\subsection{Camera estimation}
\label{supp:camera}
The input camera relative to the generated shape is estimated from the image
and the generated $512^3$ mesh, without camera supervision. A camera looking
at the origin from direction (yaw, pitch) at the training-render distance,
with free roll, field of view, and a 2D similarity on the conditioning crop,
is fitted in two steps:
(i) a grid search followed by Powell refinement of the silhouette IoU between
the input alpha mask and the projected mesh, and
(ii) an appearance-aware refinement that minimizes
$(1-\mathrm{IoU})+\lambda\,(1-\mathrm{NCC})$, where NCC compares Sobel edge
magnitudes of the input crop and of a shaded render of the mesh at
$512$\,px. Step (ii) also evaluates the mirrored candidates
(pitch $\to-$pitch, yaw $\to$ yaw$+180^\circ$, and both), which resolves the
front/back ambiguity of silhouettes. The same camera is reused at all levels.
The injected image is the conditioning crop at its native resolution
(at most $4096$\,px), masked by the dilated silhouette of the generated mesh.

\subsection{Choice of the $512^3$ start}
\label{supp:start}
\begin{figure}[t]
\centering
\includegraphics[width=\linewidth]{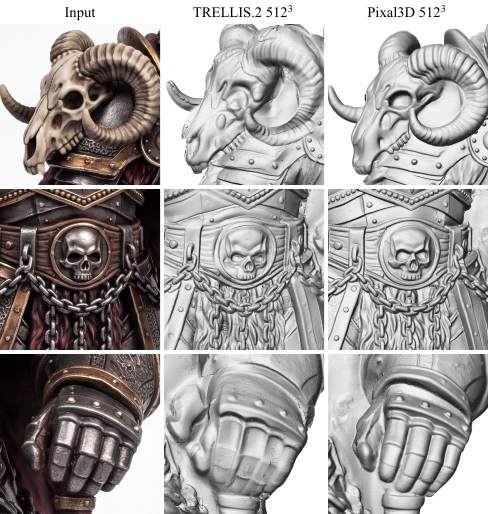}
\caption{\textbf{Inaccurate geometry from a misaligned start.} Close-ups of a generated
image (rows: shoulder, belt, gauntlet). Left to right: input image and
input-view renders of SuperFine3D started from the TRELLIS.2 $512^3$ shape and from
the Pixal3D $512^3$ shape, with all other settings identical. Where the TRELLIS.2
layout disagrees with the image, detail blurs: the shoulder skull loses its eye
socket, the buckle skull an eye, and the fingers fuse.}
\label{fig:supp_start_seams}
\end{figure}

Injection and refinement move the surface by a few voxels per level, so the
global layout, which CD-L1 and F-score measure, is that of the $512^3$ start,
and where that layout disagrees with the image, detail comes out blurred or
wrong. A $512^3$ start leaves mid-scale structure to the image, whereas the
$1024^3$ stage of TRELLIS.2 adds mid-scale structure that the image does not show.
SuperFine3D therefore starts from a pixel-aligned shape: the $512^3$ stage of
Pixal3D~\cite{li2026pixal3d}, which projects image features into the volume
through an estimated camera, so parts appear where the image shows them.
Pixal3D builds on TRELLIS.2 and shares its shape autoencoder; its $512^3$
shape, obtained in about 30\,s including model loading, is placed in the
TRELLIS.2 frame with the fitted cameras of Sec.~\ref{supp:camera} and encoded
at $512^3$. After refinement it leaves fewer floating fragments on thin parts
than the full $1536^3$ Pixal3D output. On three photographs it fits the input
with a silhouette IoU of 0.93--0.95, against 0.87--0.90 for TRELLIS.2 and
Hunyuan3D-2.1~\cite{hunyuan3d2025hunyuan3d21} starts (edge NCC 0.34--0.39
against 0.18--0.23 on the first), and under identical settings it avoids the
blurred and missing parts of a TRELLIS.2 $512^3$ start
(Fig.~\ref{fig:supp_start_seams}).

All results of the main paper start from this latent: the two-view result of
its Fig.~6 uses the two-view Pixal3D generator
(Sec.~\ref{supp:multiview}), TRELLIS.2-Recursive starts from the same shape
(Sec.~\ref{supp:baselines}), and the Pixal3D column of
its Fig.~5 is its full $1536^3$ output. The
GT@512 row of its Table~1 replaces the start by the
reference mesh, placed with its own image-only fitted camera and encoded at
$512^3$ in the same way, followed by the unchanged pipeline; it separates the
start from the refinement.

\subsection{Spatial routing and cross-scale conditioning}
\label{supp:routing}
HR cells are defined in latent-grid coordinates. Within a cell, attention
includes every active token. A bounded additional set $\mathcal N_B$ of at most $B$ tokens is selected from the
26 neighboring cells, prioritizing tokens near face-adjacent boundaries and
then spatially distributed neighbors. Routing is computed from coordinates
and reused within the stage because the denoising support remains fixed.

For $N_H$ tokens and cell occupancies $n_c$, the number of attention
interactions is bounded by
\begin{equation}
 \sum_c n_c(n_c+B)\leq N_H(b^3+B).
\end{equation}
For fixed $b$ and $B$, the HR attention interaction count grows linearly
with the active HR support. Global context is concentrated in the smaller
LR stream, while the HR stream develops detail through bounded local
interactions.

The LR stream is conditioned on the image and globally contextualized before
spatial readout by HR cells. HR and LR attention use separate softmax
normalizations. Center-aligned coordinate mapping places HR queries and LR
keys in a common coordinate frame. For trilinear anchoring, missing sparse
corners contribute zero and the remaining weights are not renormalized.
Thus, the coverage $c_i$ attenuates anchors near incomplete LR support.
The object-space structural field provides a complementary fixed spatial
footprint as the HR lattice becomes finer. Its bounded affine modulation is
zero-initialized at the output, preserving pretrained activations at
initialization. The continuous anchor and LR attention readout are fused
through the two gates defined in the main paper.

\subsection{Caching and memory management}
Within a refinement stage, the LR latent and image condition are fixed.
The LR stream uses $t_L=0$ and is evaluated independently of the evolving
HR denoising state, so its encoded context and layerwise LR keys and values are cached for spatially indexed HR access; the keys and values are kept in pinned host memory and copied to the GPU one layer ahead of use on a separate stream.
Time-dependent gates and HR activations are evaluated at every step.
 Memory-bounded decoding and rendering retain the
complete geometry at the requested resolution.

\subsection{Index-based attention and fused kernels}
\label{supp:kernels}
\paragraph{Index-based attention.}
HR tokens are sorted by cell, so the queries of a cell form one
variable-length sequence whose keys are the cell's own tokens and its
neighbor set $\mathcal N_B$ (Sec.~\ref{supp:routing}). We extend the
variable-length FlashAttention-2~\cite{dao2024flashattention2} kernel to
load key and value rows through a per-sequence array of row indices into
the shared K/V tensors.
All cells are processed in one launch, and no per-cell copy of keys and
values is formed. The HR-to-LR read uses the same kernel with indices into
the cached LR keys and values. The output is bit-identical to gathering
the rows and calling the original kernel.

\paragraph{Fused kernels.}
Hand-written Triton~\cite{tillet2019triton} kernels fuse (i) the per-head
RMS normalization~\cite{zhang2019rmsnorm} of queries and keys with the 3D
rotary embedding~\cite{su2024roformer}, (ii) LayerNorm~\cite{ba2016layernorm}
with fp32 statistics and the adaLN~\cite{peebles2023dit} scale and shift,
(iii) the gated residual updates, and (iv) the query and key normalization
of the image cross-attention; the LoRA~\cite{hu2022lora} updates are merged
into the attention projections. The kernels accumulate in fp32, round to bf16
where the reference implementation stores bf16 tensors, and use 64-bit
offsets, since element offsets exceed $2^{31}$ on the largest supports.
The kernels compile once and are cached on disk, so a stage with a single
network evaluation pays no compilation cost.

\section{Detail-Normal Estimation and Height-Field Injection}
\label{supp:injection}
\subsection{Detail-normal estimator}
The estimator is the one-step diffusion normal predictor NiRNE of
Hi3DGen~\cite{ye2025hi3dgen} (VAE, ControlNet~\cite{zhang2023controlnet}, and UNet), fine-tuned so that
it resolves pixel-scale detail.

\paragraph{Training.}
UNet and ControlNet are trained (fp32 master weights, bf16 autocast); the VAE
is frozen and gradients pass through its decoder. Each sample is a region of
$768/U$ pixels with $U\sim\log\mathcal U[2,4]$ cut from a $1024$\,px close-up
render of a training asset (camera zoomed $2\times$ or $4\times$ onto a local region,
so fine relief spans many pixels) and upsampled to the native $768$\,px processing
size, as the input is at inference;
the prediction is area-downsampled back and compared with the rendered
ground-truth normals of the region. On covered pixels the loss is
\begin{equation}
 \mathcal L_{\mathrm{est}}=\big\langle 1-\widehat{\mathbf n}^\top\mathbf n\big\rangle
 +\lambda_{\mathrm{hp}}\sum_{k\in\{4,8\}}
 \big\langle\|\mathrm{hp}_k(\widehat{\mathbf n})-\mathrm{hp}_k(\mathbf n)\|^2\big\rangle,
\end{equation}
where $\mathrm{hp}_k$ is a masked box high-pass of $k$ pixels and
$\lambda_{\mathrm{hp}}=3$. The default estimator is trained for 4,000 steps
(batch size four, learning rate $10^{-5}$) on 1,961 grey shaded close-up renders of
480 assets selected from the training set by their HR/LR token ratio; none
overlaps the evaluation assets.

\begin{figure*}[t]
\centering
\includegraphics[width=\linewidth]{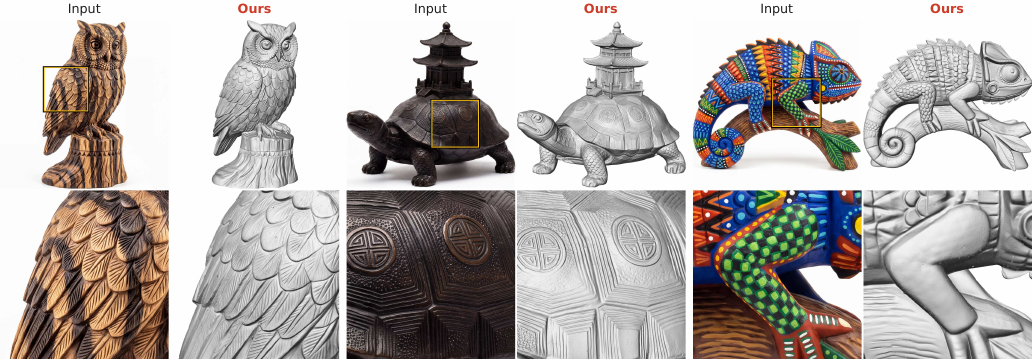}
\caption{\textbf{Image appearance is not carved into the surface.} Generated
images of a zebrawood owl with bold grain stripes, a bronze tortoise with a
near-black patina under low-key light, and a folk-art chameleon whose flat paint
runs across the carving, each with SuperFine3D at $4096^3$ from the input view
(top) and close-ups of the boxed region (bottom). The carved feathers, chased
medallions, and carved scales are recovered; the grain stripes and the painted
checks do not become relief.}
\label{fig:supp_textured}
\end{figure*}

\paragraph{Textured inputs.}
On textured photographs, relief must be separated from albedo and cast
shadows. We therefore condition the estimator on the current mesh: the UNet
starts from the VAE latent of the mesh's rendered normals, while the image
drives the ControlNet, so appearance changes without geometric support are
not turned into relief (Fig.~\ref{fig:supp_textured}). This variant is fine-tuned from the default
estimator with the same loss and settings, first on textured renders under
two lighting setups and then with the mesh conditioning, whose normal map is
either the render of the $512^3$ shape or the reference normals low-passed
at $8$--$64$\,px. It is trained
on a mix of shaded and textured, lit renders, with an additional set of
relief and scanned assets (2,823 views of 687 assets in total; near-flat
crops removed; none overlaps the evaluation or calibration assets), and with augmentations that model the inference
conditions: misalignment of the conditioning normals (shift, rotation,
scale, smooth warp, dropout), photometric jitter, smooth random albedo fields, and
exposure normalization. Inputs whose mean foreground chroma exceeds one grey
level are routed to this variant.

\paragraph{Tiled inference.}
The input crop is upsampled $3\times$ (Lanczos) and predicted in overlapping
$768$\,px tiles with stride $3/4$ of the tile, blended with a separable tent
window, renormalized, and area-downsampled to the input resolution.
Predictions are mapped from the camera frame to the object frame with the
estimated camera.

\subsection{Band-limited normal target}
Let $\mathbf N_M$ be the normals of the current mesh rendered under the
estimated camera and $\widehat{\mathbf N}$ the estimate on the same pixel
grid. With a masked box low-pass $L_\kappa$ of $\kappa$ pixels, the target is
\begin{equation}
 \begin{aligned}
 \mathbf T=\operatorname{normalize}\Big(&L_\kappa(\mathbf N_M)
 +g\big(\widehat{\mathbf N}-L_\kappa(\widehat{\mathbf N})\big)\\
 &+(g_f-g)\big(\widehat{\mathbf N}-L_{\kappa_f}(\widehat{\mathbf N})\big)\Big).
 \end{aligned}
\end{equation}
The generated shape supplies all structure coarser than $\kappa=64$\,px and
the image supplies the detail above it; $\kappa_f=4$\,px separates the finest
band. With $g_f=g$ and without the evidence gate below, $\mathbf T$ reduces to the
single-band target $\operatorname{normalize}\big(L_\kappa(\mathbf N_M)+g(\widehat{\mathbf N}-L_\kappa(\widehat{\mathbf N}))\big)$.

\paragraph{Evidence gate.}
The gate keeps dense low-amplitude texture of the estimate out of the geometry.
Let $\mathbf d=\widehat{\mathbf N}-L_\kappa(\widehat{\mathbf N})$ be the image detail,
$e$ its root-mean-square magnitude over $4$\,px, and
$s=\operatorname{clamp}(1-\tau/e,0,1)$ a soft threshold at $\tau=2\sin(\theta/2)$
for a threshold angle $\theta$. The two detail terms of $\mathbf T$ are weighted by
$s$, and the mesh keeps its own detail $\mathbf N_M-L_\kappa(\mathbf N_M)$ with
weight $1-s$: strong image features pass almost unchanged, and the generated
detail remains wherever the image shows none.

\paragraph{Calibrated detail gains.}
A regression-trained estimator returns the conditional mean of the detail
and therefore represents its amplitude in a systematically reduced form. We
calibrate the gains once, on 13 held-out assets that appear in no training
set, figure, or table: for each asset, the estimator is applied exactly as at
inference to a $4096$\,px render, and its band-passed normals are compared
with the rendered reference normals. The amplitude ratio is remarkably
stable across assets (Table~\ref{tab:supp_calib}). We set $g=1.55$, the median
ratio for $4$--$64$\,px, so the injected relief reaches the reference depth,
and $g_f=1.2$, the least-squares gain $\bar\rho\,\sigma_{\mathrm{ref}}/
\sigma_{\mathrm{est}}$ for the finest band, whose lower correlation
$\bar\rho$ calls for a conservative gain. The same gains are used at every
level. By default, textured inputs use $g\le1.1$ with a single detail band; the settings of the figures and of the quantitative evaluation are given in Sec.~\ref{supp:settings}.

\begin{table}[t]
\centering\small\setlength{\tabcolsep}{5pt}
\caption{\textbf{Estimator calibration} on 13 held-out assets: ratio of
reference to estimated detail amplitude (median and interquartile range) and
correlation of the band-passed normals.}
\label{tab:supp_calib}
\begin{tabular}{@{}lccc@{}}
\toprule
Band (px) & Ratio & IQR & Corr.\\
\midrule
$0$--$4$   & 1.90 & 1.82--2.00 & 0.64\\
$4$--$16$  & 1.57 & 1.53--1.62 & 0.87\\
$16$--$64$ & 1.55 & 1.49--1.58 & 0.91\\
\bottomrule
\end{tabular}
\end{table}

\subsection{Height-field fitting}
Each vertex $\mathbf v$ visible from the camera (depth test with tolerance
$2/1024$ object units) moves along the viewing direction $\boldsymbol\omega$
by a displacement read from a height field $s$ on the image grid:
\begin{equation}
 \mathbf v'=\mathbf v+h_R\,\gamma
 \tanh\!\big(s(\pi(\mathbf v))/\gamma\big)\,\boldsymbol\omega,
\end{equation}
with $h_R=1/R$ the voxel width at level $R$ and $\gamma=6$ voxels.
The field minimizes, over target pixels covered by the mesh,
\begin{equation}
 \big\langle 1-\mathbf n_q^\top\mathbf T_q\big\rangle
 +\alpha\big\langle w_q\,\rho_{0.05}(\nabla^2 s_q)\big\rangle
 +\beta\langle s_q^2\rangle,
\end{equation}
where $\mathbf n_q$ is the normal of the displaced surface at pixel $q$,
$\rho_{0.05}$ a Huber penalty with threshold $0.05$ voxels,
$\alpha=0.003$, and $\beta=10^{-3}$. The weight
$w_q=1/(1+(\|\nabla I_q\|/(0.5\,Q_{0.9}))^2)$ relaxes smoothing on image
edges, with $Q_{0.9}$ the 90th percentile of the grey-image gradient
magnitude on the foreground. We run Adam~\cite{kingma2015adam} (learning rate $0.1$, 800
iterations) coarse-to-fine over three grid levels, re-rasterizing the
pixel-to-face assignment every 25 iterations. The bound $\gamma$ ties the
injected amplitude to the current voxel size, so each level adds relief at
its own scale and preserves the global layout by construction.

\paragraph{Resolution gate.}
\label{supp:gate}
An injection can only transfer detail that the image resolves at the scale of
the current level. Before each injection, we therefore measure how many
input-image pixels one voxel of the level spans, at the object centre along
the camera's horizontal axis. Each level halves the voxel and thus this
footprint; for the $4096$\,px input of Fig.~\ref{fig:supp_input_res} it is
8.4, 4.2, 2.1, and 1.05 pixels at $512^3$, $1024^3$, $2048^3$, and $4096^3$.
Below 0.5 pixels per voxel, more than two voxels share one pixel, so the image
holds no information at the scale that the level adds, and fitting the height
field would only imprint pixel-level estimator noise. Such a level is not
injected; its detail comes from refinement alone, while all coarser levels
are injected as usual.

\subsection{Input resolution}
\label{supp:input_res}
\begin{figure*}[t]
\centering
\includegraphics[width=\linewidth]{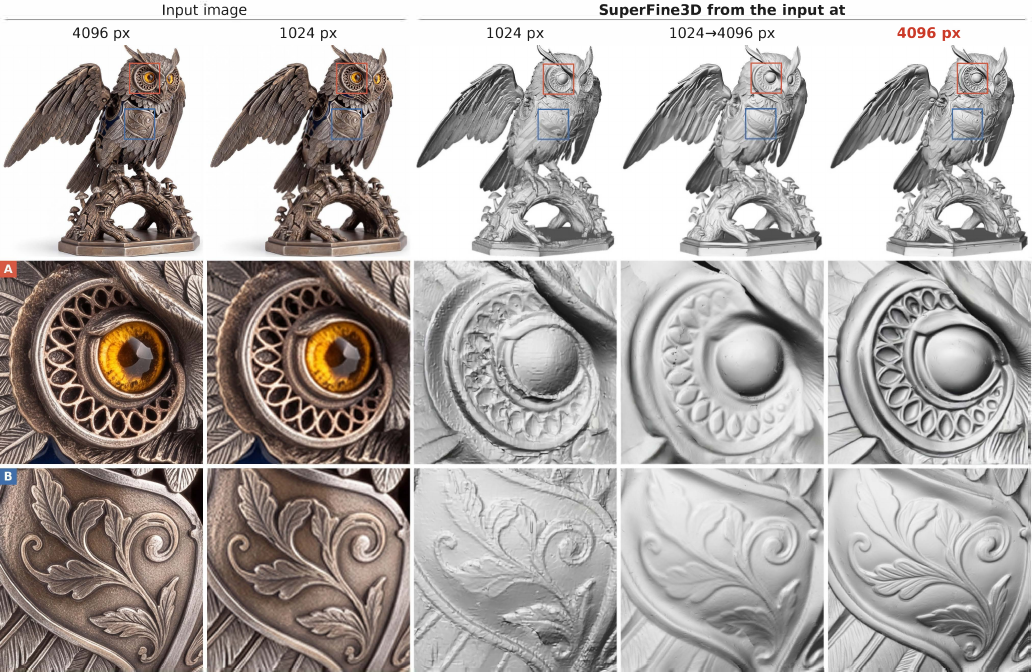}
\caption{\textbf{Detail follows the genuine image resolution.} Three versions
of one generated image, each refined by SuperFine3D from the same $512^3$
start with identical settings: the native $4096$\,px image (a genuinely
high-resolution input), the image downsampled to $1024$\,px (a genuinely
low-resolution input), and the $1024$\,px image upsampled to $4096$\,px by
Lanczos interpolation (a low-resolution image enlarged to a high pixel count).
Input-view renders with close-ups A and B. Fine relief requires genuine
high-frequency detail in the image; interpolation adds pixels but no detail.}
\label{fig:supp_input_res}
\end{figure*}

Injection transfers detail that is present in the input image, so the finest
relief SuperFine3D forms is set by the detail the image genuinely resolves,
not by its pixel count. Figure~\ref{fig:supp_input_res} isolates this with
three versions of one generated image, all refined from the same $512^3$
start with identical settings:
(i) the native $4096$\,px image, which stands for a genuinely high-resolution
input such as a high-resolution photograph or a native $4096$\,px generation;
(ii) the image downsampled to $1024$\,px, which stands for a genuinely
low-resolution input such as a typical $1024$\,px web image or generator
output; and
(iii) that $1024$\,px image upsampled back to $4096$\,px by Lanczos
interpolation, which stands for a low-resolution image enlarged to a high
pixel count: it has the pixel count of (i) but only the detail of (ii).

With input (i), SuperFine3D forms the openwork eye rim and the leaf scrolls of
the chest plate (A, B). With input (ii), these structures appear only in
coarse form: the image does not resolve them, and the resolution gate
(Sec.~\ref{supp:gate}) injects at $512^3$ to $2048^3$ but skips the final
$4096^3$ injection, where one voxel spans only 0.26 pixels. Input (iii) has
four times as many pixels and passes the gate at every level (1.05 pixels per
voxel at $4096^3$), but the gate measures the pixel footprint, not the detail:
its pixels carry no detail beyond that of (ii), every injection transfers the
smooth interpolated normals, and the relief comes out rounded. Genuine resolution thus translates directly into geometric detail,
whereas interpolation adds pixels but no detail.

\paragraph{Practical note.}
High-resolution results therefore require an input image with genuine
high-frequency detail; a high pixel count alone is not sufficient. Images
should be used at the resolution at which they were captured or generated,
and not upsampled by interpolation beforehand. A simple spectral check
separates the two cases: the share of spectral power above the Nyquist
frequency of a $2048$\,px image is 3.0\% for input (i) and 0.007\% for
input (iii); all inputs of Fig.~5 lie at
1.2--3.0\%.

\subsection{Settings of figures and evaluation}
\label{supp:settings}
Figures~1, 5, 6, and~\ref{fig:supp_input_res} use $t_0=0.05$ at the $4096^3$ stage.
In these figures, textured inputs use $g=1.5$ and keep the finest
band ($g_f=1.2$, $\kappa_f=4$\,px). Before the target is
formed, the estimate is warped onto the rendered mesh normals by a
smooth dense flow (DIS optical flow~\cite{kroeger2016dis} on low-passed normal images,
Gaussian-smoothed and bounded to 96\,px), so image detail lands on
the mesh's own structures. Figures~1, 5, and~\ref{fig:supp_input_res} also average the estimate
over two tile offsets and a horizontal mirror.

\paragraph{Quantitative evaluation.}
Tables~1, \ref{tab:supp_imgfid}, and~\ref{tab:supp_offview} use one setting for
all inputs: $t_0=0.05$ at the $4096^3$ stage, unit gains ($g=g_f=1$), and the
evidence gate with $\theta=8^\circ$ ($16^\circ$ at the final $4096^3$ injection).
Each finer level injects only finer bands, with $\kappa=64$\,px at $512^3$ and
$1024^3$, $32$\,px at $2048^3$, and $16$\,px at the final $4096^3$ injection.
The estimate is computed once at the first level and reused at all levels,
without flow warp or test-time averaging.

\section{Multi-View Input}
\label{supp:multiview}
For the two-view experiment (Fig.~6), cameras are recovered from the photos (Cleveland Museum of Art Open Access, CC0): dense MASt3R~\cite{mast3r} correspondences give the relative pose through the essential matrix, with the focal length of the MASt3R pointmaps, and the scale follows from the first view's generated depth. Each photo is warped by a rotation homography to a camera that looks at the object centre. The multi-view coarse generator of Pixal3D~\cite{li2026pixal3d} averages the pixel-aligned features of both views; the single-view result uses the same generator with the first view only. The resulting $512^3$ shape enters the unchanged pipeline, and detail injection uses the first view. The two inputs are $50^\circ$ apart. Held-out photos are posed the same way.

\section{Geometric Supervision}
\label{supp:losses}
\subsection{Frozen-decoder targets and valid observations}
The geometric objectives use surfaces decoded from paired target LR and HR
latents, i.e.\ frozen-decoder reconstructions of training targets.
Benchmark evaluation uses the original reference triangles. Decoder weights
are frozen, while gradients propagate through the predicted branch to the
refinement model. Target features, target surfaces, reliability masks, and
correspondence decisions are detached. The objectives are evaluated on the
clean estimate for $t\le0.5$, every fourth optimization step, and are added to flow matching~\cite{lipman2023flow} with weights
$\lambda_{\mathrm k}=0.0117$, $\lambda_{\mathrm d}=0.0282$,
$\lambda_{\mathrm e}=0.0087$, and $\lambda_{\mathrm p}=0.0111$.

Geometric supervision uses full-context latent decoding, followed by local
mesh extraction and rendering for loss evaluation. Teacher and prediction
branches use independent sparse-tensor caches to preserve the predicted
branch's differentiable computation. Local supervision excludes the
extraction boundary. The implementation uses opposing views of the same
local coordinate frame.

Let $v_L$ and $v_H$ denote target visibility masks. A target reliability mask is
\begin{equation}
 \begin{aligned}
 m_n&=\mathbf{1}[\mathbf n_L^\top\mathbf n_H>\gamma_n],\\
 m_d&=\mathbf{1}[|d_H-d_L|<\gamma_d h_L],\\
 m_T&=v_Lv_Hm_nm_d,
 \end{aligned}
\end{equation}
with $\gamma_n=0.8$ and $\gamma_d=4$.
The mask is eroded by one pixel before finite differences are evaluated.
With $E(m_T)$ this erosion and $v_P$ the prediction visibility, the mask used
for rendered correction losses is
\begin{equation}
 m=\operatorname{stopgrad}(E(m_T)v_P).
\end{equation}

\subsection{Signed corrections and finite differences}
The normalized corrections are
$\delta^\star=(d_H-d_L)/h_L$ and
$\widehat\delta=(\widehat d_H-d_L)/h_L$,
with all depths and $h_L$ in the same local physical frame.
We use $\tau=0.5$ and $\eta=0.25$ LR voxels for the preservation/detail
weights and the preservation tolerance. With the smooth-$L_1$ penalty
\begin{equation}
 \rho_\beta(e)=
 \begin{cases}e^2/(2\beta),& |e|<\beta,\\
 |e|-\beta/2,&\text{otherwise},
 \end{cases}
\end{equation}
the signed correction term is
$\mathcal L_{\mathrm{signed}}=
\langle\rho_{0.25}(\widehat\delta-\delta^\star)\rangle_{w_{\mathrm{detail}}}$.
For image-plane direction $e\in\{x,y\}$, with forward differences
$D_e\delta(p)=\delta(p+e)-\delta(p)$ and valid pair weights
$w_e(p)=w_{\mathrm{detail}}(p)m(p+e)$,
\begin{equation}
 \mathcal L_\nabla=\frac12\sum_{e\in\{x,y\}}
 \big\langle\rho_{0.25}(D_e\widehat\delta-D_e\delta^\star)\big\rangle_{w_e}.
\end{equation}
The normal term is
$\mathcal L_{\mathrm{normal}}=\langle 1-\operatorname{clamp}
(\widehat{\mathbf n}_H^\top\mathbf n_H,-1,1)\rangle_{m_N}$,
with detached shared visibility and target reliability $m_N$ and
$\lambda_n=0.1$. View losses are averaged. Rendered losses are disabled when
fewer than 64 reliable pixels remain across the selected views;
decoder-space supervision remains available in that case.

\subsection{Balanced surface-crossing supervision}
Only coordinates shared by predicted and target decoder outputs receive surface-crossing labels; an unmatched coordinate is unknown rather than a
negative example. For axis $a\in\{1,2,3\}$ and label $c\in\{0,1\}$, let
$\mathcal M_{a,c}=\{i\in\mathcal M:y_{i,a}=c\}$. Then
\begin{equation}
 \begin{aligned}
 L_{a,c}&=\frac{1}{|\mathcal M_{a,c}|}
 \sum_{i\in\mathcal M_{a,c}}\rho_{\mathrm{BCE}}(\ell_{i,a},c),\\
 \mathcal L_{\mathrm{edge}}&=\frac13\sum_{a=1}^3
 \frac{1}{|\mathcal C_a|}\sum_{c\in\mathcal C_a}L_{a,c},
 \end{aligned}
\end{equation}
where $\rho_{\mathrm{BCE}}$ is binary cross-entropy with logits and
$\mathcal C_a$ contains the nonempty classes. Empty matched sets contribute
zero, and the larger class cannot dominate the sparse prediction task.

\subsection{Reliable tangent-plane correspondences}
Predicted and target vertex offsets are compared at matched decoder
coordinates. Target normals are constructed only where decoded target
positions coincide with teacher mesh vertices (tolerance $0.02$ HR voxels),
from at least three nondegenerate, sign-aligned incident faces whose minimum
absolute normal agreement exceeds $0.9$; sharp or inconsistent neighborhoods
are excluded. Positions are expressed in HR voxel units and $\rho_{0.1}$ is
applied to the normal component of the displacement only, which preserves
tangential flexibility; the penalty is invariant to the sign of the target
normal.

\subsection{Conditional flow matching}
For target HR features $\mathbf f_0^H$, Gaussian noise $\epsilon$, and
$t\sim\mathcal U(0,1)$,
\begin{equation}
    \mathbf f_t^H=(1-t)\mathbf f_0^H+\sigma_t\epsilon,
    \qquad \sigma_t=\sigma_{\min}+(1-\sigma_{\min})t,
\end{equation}
with $\sigma_{\min}=10^{-5}$. The target velocity, objective, and clean
estimate are
\begin{equation}
    \begin{aligned}
        \mathbf v^\star&=(1-\sigma_{\min})\epsilon-\mathbf f_0^H,\\
        \mathcal L_{\mathrm{FM}}&=
        \mathbb E\big[\|v_\theta(\mathbf f_t^H,t,I,z^L)-\mathbf v^\star\|_2^2\big],\\
        \widehat{\mathbf f}_0^H&=(1-\sigma_{\min})\mathbf f_t^H-\sigma_t v_\theta .
    \end{aligned}
    \label{eq:flow_loss}
\end{equation}
The image condition is dropped with probability $0.1$.

\section{Training and Reproducibility Settings}
\label{supp:training}
\begin{table}[t]
\centering\small\setlength{\tabcolsep}{4pt}
\caption{\textbf{Training settings of the refinement operator.}}
\label{tab:supp_train}
\begin{tabular}{@{}ll@{}}
\toprule
Data & 10,000 assets, $512^3$--$1024^3$ latent pairs\\
Initialization & TRELLIS.2 $1024^3$ shape flow model\\
Stage 1 & HR attention LoRA only, 5,000 steps\\
Stage 2 & + LR conditioning, 2,000 steps\\
LoRA & rank 16, all 30 blocks, lr $\times0.5$ in Stage 2\\
Optimizer & AdamW~\cite{loshchilov2019decoupled}, lr $2\times10^{-5}$, wd $0.01$\\
Batch & 2 per GPU, bf16\\
LR perturbation & $p=0.5$, token deletion, noise $0.1$\\
Image encoder & DINOv3~\cite{simeoni2025dinov3} ViT-L/16, $1024$\,px\\
\bottomrule
\end{tabular}
\end{table}
Table~\ref{tab:supp_train} lists the refinement training settings. All other
pretrained parameters, the geometry autoencoder, and the image encoder are
frozen. The random LR perturbation makes it robust to the generated LR
latents it receives at inference. The accompanying configuration records
routing settings, loss weights, asset identifiers, preprocessing hashes, and
split membership.

\begin{table*}[t]
\centering\small\setlength{\tabcolsep}{8pt}
\caption{\textbf{Image-space normal fidelity on rendered relief assets}: correlation of
$4$--$16$\,px / $16$--$64$\,px band-passed normals with the reference ($\uparrow$).
The first four inputs are untextured renders, the last a textured render of the door.}
\label{tab:supp_imgfid}
\begin{tabular}{@{}lccccc@{}}
\toprule
Method & Censer & Door & Triptych & Dragon & Door (tex.)\\
\midrule
TRELLIS.2            & 0.00 / 0.00 & 0.00 / 0.01 & 0.00 / 0.04 & 0.01 / 0.08 & 0.00 / $-$0.03\\
Hunyuan3D-2.1        & 0.02 / 0.13 & 0.00 / $-$0.02 & 0.00 / 0.02 & 0.01 / 0.02 & $-$0.01 / $-$0.01\\
Pixal3D              & 0.02 / 0.14 & 0.02 / 0.24 & 0.00 / 0.01 & 0.04 / 0.22 & 0.02 / 0.22\\
Direct3D-S2          & 0.00 / $-$0.01 & 0.00 / 0.03 & 0.00 / 0.01 & 0.01 / 0.03 & 0.00 / $-$0.02\\
SAM 3D               & 0.00 / 0.05 & $-$0.01 / 0.25 & 0.00 / 0.02 & 0.00 / 0.02 & 0.02 / 0.06\\
UltraShape           & 0.00 / 0.06 & 0.01 / 0.25 & 0.00 / 0.04 & 0.00 / 0.03 & 0.01 / 0.00\\
TRELLIS.2--Recursive & 0.02 / 0.17 & $-$0.01 / 0.09 & 0.01 / 0.09 & 0.00 / 0.10 & 0.00 / 0.13\\
\midrule
SuperFine3D          & \textbf{0.29} / \textbf{0.51} & \textbf{0.66} / \textbf{0.55} & \textbf{0.35} / \textbf{0.59} & \textbf{0.30} / \textbf{0.48} & \textbf{0.35} / \textbf{0.41}\\
\bottomrule
\end{tabular}
\end{table*}

\section{Baseline Configurations and Comparison Scope}
\label{supp:baselines}
\paragraph{TRELLIS.2 released generator.}
The generation-quality baseline uses the official $1536$ cascade setting.

\paragraph{TRELLIS.2--Recursive.}
The recursive baseline reapplies the official generator weights on
successively finer decoder-predicted supports, beginning from the shared
$512^3$ initialization. It reaches the same $4096^3$ grid as SuperFine3D and therefore measures
what the official generator achieves from grid resolution alone. For image-to-mesh comparisons, each method follows its own
generated supports; for block-level replay, supports and tensors are held
identical.

\paragraph{Hunyuan3D-2.1 and Pixal3D.}
Hunyuan3D-2.1 uses $1024$ field extraction, an extraction density of a
continuous field rather than a native $1024^3$ diffusion latent. Pixal3D uses
the official main-release $1536$ geometry setting.

\paragraph{UltraShape.}
UltraShape~\cite{jia2025ultrashape} refines the corresponding Hunyuan3D-2.1 coarse mesh ($1024$
extraction) using the original RGB input, with its official configuration
(50 sampling steps, 32,768 latent tokens, 128 query voxels, seed 42). We use its
highest extraction setting ($4096$ requested, $4032$ cells per axis under FlashVDM~\cite{lai2025flashvdm}), extracted in memory-bounded slabs that reproduce the official
extraction exactly and keep the full-resolution mesh.

\paragraph{SAM 3D.}
SAM 3D~\cite{sam3d2025} receives the same RGBA input (the alpha channel as the
object mask) and runs with the released checkpoint and the settings of its
reference inference script (seed 42; no mesh post-processing, texture baking,
or layout optimization). We use the raw output of its mesh decoder, a
FlexiCubes surface extracted on a $256^3$ grid, without simplification.

\section{Evaluation Data and Matched Comparisons}
\label{supp:data}
The benchmark is drawn from a frozen Toys4K~\cite{stojanov2021using}
asset manifest and includes
complex shapes with thin structures, separated components, or detailed
surface relief. Every method receives the same original input image, rendered
at $4096$\,px, and is compared with the same original reference geometry,
which is used only for evaluation. For a requested asset set $\mathcal A$ and method set
$\mathcal K$, all methods in a table are averaged on the common complete cohort
\begin{equation}
 \mathcal A_{\mathrm{common}}
 =\bigcap_{k\in\mathcal K}\mathcal A_k^{\mathrm{valid}},
\end{equation}
where $\mathcal A_k^{\mathrm{valid}}$ contains the assets with valid metrics
for method $k$; the manifest records the requested set, valid results, and
coverage. Each paired ablation of the main paper fixes the checkpoints,
initialization, sampling settings, seed, decoder, and camera.

\paragraph{Generated inputs.}
The inputs of
Figs.~5, \ref{fig:supp_start_seams}, \ref{fig:supp_textured}, and~\ref{fig:supp_input_res}, and the lion and pangolin of Fig.~\ref{fig:supp_cmp_all} are images generated at $4096$\,px with Nano
Banana Pro (Gemini 3 Pro Image)~\cite{nanobananapro2025}; every method receives the same image. They
carry genuine detail at this resolution (Sec.~\ref{supp:input_res}).

\section{Geometric Evaluation}
\label{supp:metrics}
\subsection{Normalization and registration}
Each mesh is normalized with its triangle-area-weighted surface centroid and
the diameter-one sphere convention:
\begin{equation}
 \mathbf c=\frac{\sum_f A_f\mathbf c_f}{\sum_f A_f},\quad
 s=2\max_{\mathbf v\in V}\|\mathbf v-\mathbf c\|_2,\quad
 \widetilde{\mathbf v}=(\mathbf v-\mathbf c)/s .
\end{equation}
Prediction and reference are normalized independently. Each prediction is
then registered to the reference with a proper rigid transform (no scale or
reflection), which is frozen for all reference-based metrics of that
prediction. Registration samples 100,000 surface points per mesh,
deterministically retains at most 10,000, considers world-axis and PCA-frame
rotation hypotheses, refines the eight best with at most 80 rigid
nearest-neighbor iterations, and selects the smallest symmetric squared
nearest-neighbor objective. Final distances use the original triangles, with
two million area-uniform samples per surface (seeds 101 and 102).

\subsection{CD-L1 and F-score}
With prediction and reference samples $P$, $Q$ and
$d(p,M)=\min_{x\in M}\|p-x\|_2$,
\begin{equation}
 \operatorname{CD\text{-}L1}=\tfrac12\Big(\tfrac1{|P|}\textstyle\sum_{p}d(p,M_G)
 +\tfrac1{|Q|}\sum_{q}d(q,M_P)\Big),
\end{equation}
and, at threshold $\tau$ with a strict test,
$P_\tau=\frac1{|P|}\sum_p\mathbf 1[d(p,M_G)<\tau]$,
$R_\tau=\frac1{|Q|}\sum_q\mathbf 1[d(q,M_P)<\tau]$,
$F_\tau=2P_\tau R_\tau/(P_\tau+R_\tau)$ ($0$ if undefined).
Distances are Euclidean; entries labeled $\times10^3$ are multiplied by 1000.

\subsection{Detail Quality}
DQ is the MeshyBench~\cite{meshybench2026} quality\_4096
field (eight normal views
at $8192$ raster resolution, base resolution 512, stratified $8\times8$
sampling). Large meshes are rendered in batches of two million triangles with exact
per-pixel depth merging, so every triangle contributes. DQ measures reference-free detail; CD-L1 and F-score measure agreement with
the reference shape.

\subsection{Image-space normal fidelity}
\label{supp:imgmetric}
DQ measures how much detail is formed. To show that this detail follows
the input image, reference normals are ray cast with the camera that
rendered the input (orthographic for Toys4K), on the pixel grid of the crop of the $4096$\,px input; each result is
rendered with its own image-only fitted camera on the same grid, and both are
compared as camera-frame normals on pixels covered by both. We report the mean
angle between the two normal maps (Ang.) and, for the bands
$[a,b]\in\{[4,16],[16,64]\}$\,px, the
correlation between the band-passed normals $L_a(\mathbf n)-L_b(\mathbf n)$ of
result and reference. Higher correlation means the generated detail reproduces the reference
detail in that band.
Table~1 reports this measure on the Toys4K cohort and Table~\ref{tab:supp_imgfid} on five rendered relief assets with reference geometry; Sec.~\ref{supp:offview} evaluates it off the input view.

\subsection{Off-view detail fidelity}
\label{supp:offview}
Detail injection reads the input view, so we test whether the injected relief
is 3D geometry that holds from other directions. On the Toys4K cohort of
Table~1, the reference mesh is turned about its
vertical axis by $\pm30^\circ$ and $\pm60^\circ$ and rendered with the input
camera; each result is turned by the same rotation, expressed in its own frame
through its image-only fitted camera, and rendered with that camera. Surface
points are matched through the input view: a reference point visible in the
input view is paired with the result's surface on the same input ray, as in
Sec.~\ref{supp:imgmetric}, and both must be visible in the turned view. Errors of
the global shape therefore change where a point appears in the turned view, not
which points are compared. We correlate the band-passed normals of the paired
points, each computed on its own mesh's turned render, and report the amplitude
ratio of the two. At $0^\circ$ the measure is the Corr of
Table~1.

\begin{figure*}[!t]
\centering
\includegraphics[width=\linewidth]{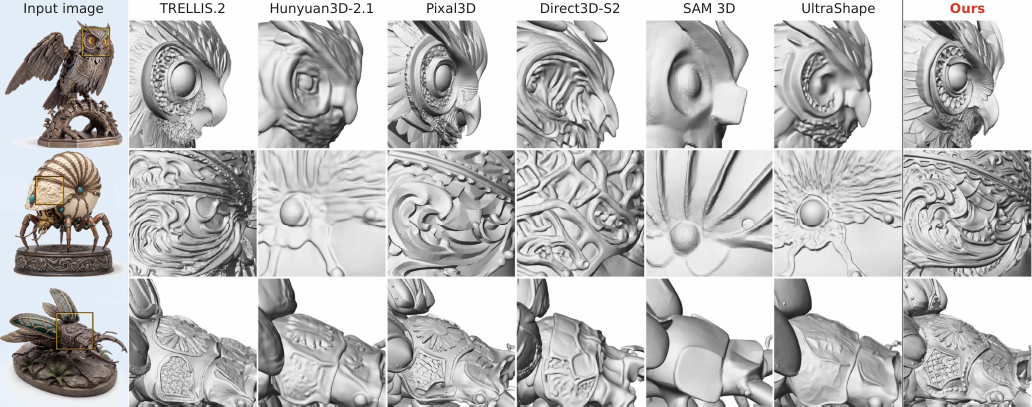}
\caption{\textbf{Detail holds off the input view.} The generated images of
Fig.~5. Each method's close-up of the boxed
region is rendered with the object turned by $30^\circ$ about the vertical axis
through that region, so every panel shows the same surface $30^\circ$ off the
input view. The relief that SuperFine3D transfers from the input image keeps its
shape and depth in the turned view.}
\label{fig:supp_offview}
\end{figure*}

Table~\ref{tab:supp_offview} reports the correlation. SuperFine3D keeps 71\% of
its $4$--$16$\,px correlation at $\pm30^\circ$ and 56\% at $\pm60^\circ$, and at
every angle stays 4.0 to 9.7 times above the best baseline, which holds no
detail in this band from any direction. Its amplitude ratio stays between 0.9
and 1.1 at all angles, so the relief keeps the reference depth from every tested
direction, not only from the input view. The decrease with angle includes the
changed sampling: surfaces seen obliquely in the input view are sampled coarsely
there, and a pixel band covers a different surface scale in the turned view.
Figure~\ref{fig:supp_offview} shows the close-ups of
Fig.~5 seen $30^\circ$ off the input view.

\begin{table}[t]
\centering\small\setlength{\tabcolsep}{4pt}
\caption{\textbf{Off-view detail fidelity} on the Toys4K cohort: correlation of
band-passed normals with the reference at surface points matched through the
input view, with the object turned by $0^\circ$, $\pm30^\circ$, and $\pm60^\circ$
(mean of both sides) ($\uparrow$). $^\dagger$On its 20-asset subset.}
\label{tab:supp_offview}
\begin{tabular}{@{}lcccccc@{}}
\toprule
 & \multicolumn{3}{c}{$4$--$16$\,px} & \multicolumn{3}{c}{$16$--$64$\,px}\\
\cmidrule(lr){2-4}\cmidrule(l){5-7}
Method & $0^\circ$ & $\pm30^\circ$ & $\pm60^\circ$ & $0^\circ$ & $\pm30^\circ$ & $\pm60^\circ$\\
\midrule
TRELLIS.2            & 0.002 & 0.014 & 0.024 & 0.105 & 0.117 & 0.149\\
Hunyuan3D-2.1        & 0.006 & 0.020 & 0.027 & 0.110 & 0.119 & 0.126\\
Pixal3D              & 0.038 & 0.044 & 0.050 & 0.268 & 0.254 & 0.251\\
UltraShape$^\dagger$ & 0.014 & 0.019 & 0.026 & 0.166 & 0.154 & 0.170\\
\midrule
SuperFine3D          & \textbf{0.368} & \textbf{0.259} & \textbf{0.204} & \textbf{0.518} & \textbf{0.401} & \textbf{0.358}\\
\bottomrule
\end{tabular}
\end{table}

\subsection{Additional comparisons}
\label{supp:gallery}
\begin{figure*}[!t]
\centering
\includegraphics[width=\linewidth]{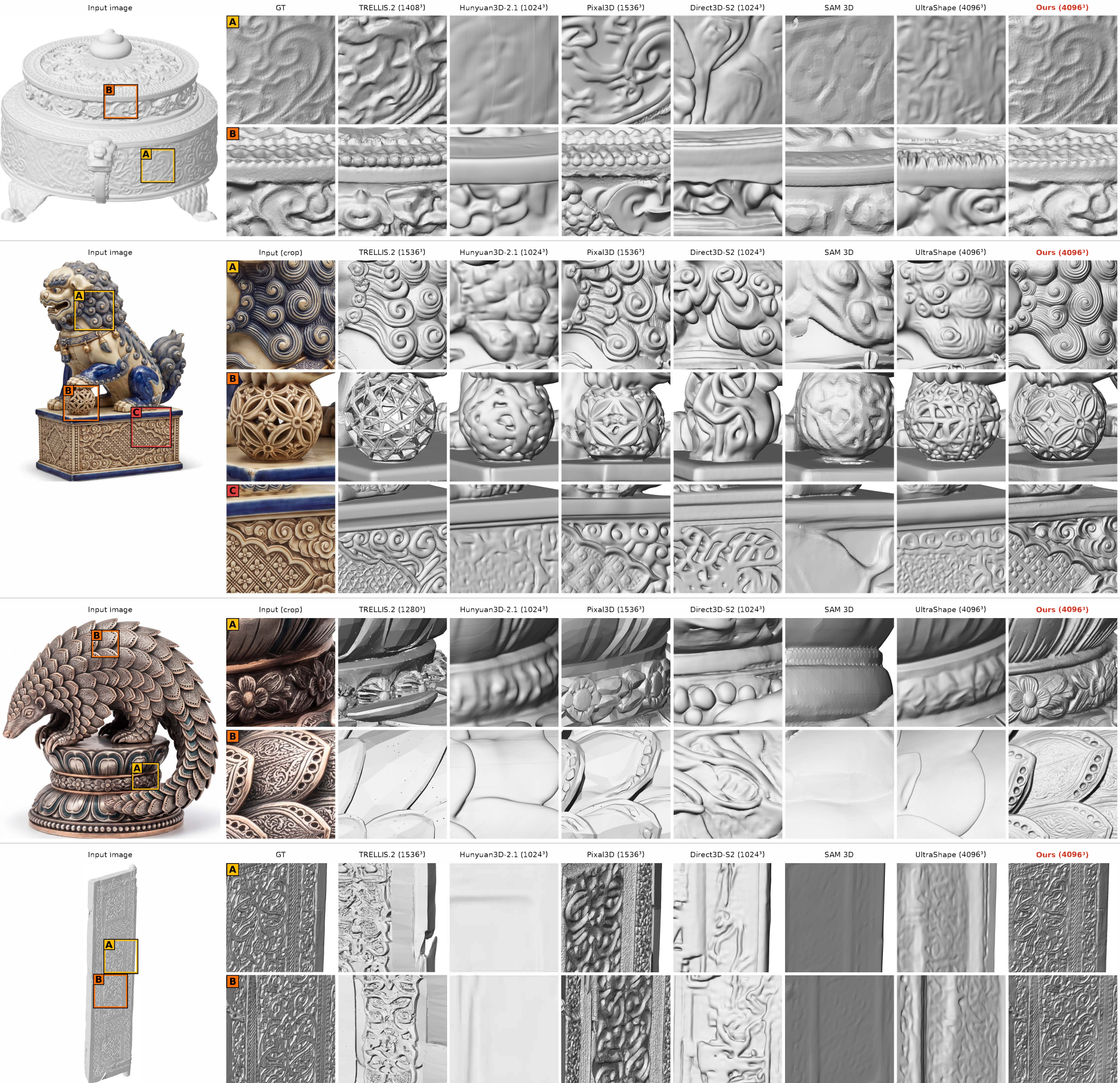}
\caption{\textbf{Additional comparisons with existing methods.} From top to
bottom: carved censer, guardian lion and pangolin (generated images), and
ornate wooden door. Left: input image with regions A--C;
right: close-ups of the reference (for generated images, the input crop) and of each
method, rendered with identical material and lighting from the input view.
TRELLIS.2 falls back to $1408^3$ on the censer and to $1280^3$ on the pangolin.
}
\label{fig:supp_cmp_all}
\end{figure*}

Figure~\ref{fig:supp_cmp_all} compares SuperFine3D with the baselines on two
relief-rich scanned assets rendered at $4096$\,px and two generated
$4096$\,px images. SuperFine3D reproduces the carved and sculpted detail of
the input at its positions, whereas the baselines either smooth it away or
replace it with patterns that do not match the input;
Table~\ref{tab:supp_imgfid} quantifies this on the rendered relief assets,
including the censer and the door.

\section{Efficiency Analysis}
\label{supp:efficiency}
SuperFine3D reaches a $4096^3$ decoder grid in 176.2\,s from a single image
(Table~2). Three properties of the design produce this
speed: each refinement stage needs one evaluation of the refinement network
on a sparse support; spatial-cell attention runs as index-based kernels that
never copy neighbor keys and values; and fused kernels merge the
memory-bound operators of each transformer block. This section quantifies
each of them.

\subsection{Measurement protocol}
\label{supp:eff_protocol}
All measurements use one NVIDIA RTX PRO 6000 Blackwell (Max-Q) GPU.
A measurement starts only on an idle GPU: no other compute process,
utilization at most 2\% and less than 1.5\,GiB in use over three checks
30\,s apart, and host load below half of the CPU cores. During the
measurement the GPU is polled every second, and a run during which any
other process appears is discarded and repeated. Stage boundaries are
CUDA-synchronized. End-to-end runs start from the input image and end with
the exported mesh, including model loads, host--device transfers, and file
I/O, and excluding interpreter start-up and imports. Helper processes (the
coarse-shape generator and the voxelization workers) are started before the
measurement, so the same exclusion applies to them; all of their work,
including model loads, is measured. Kernel and block
measurements report the median of 15 and 9 synchronized repetitions after
warm-up; block measurements, including those in Fig.~1, replay
transformer block 0 (conditional branch, batch size one) on matched supports. Memory is the peak allocated GPU memory.

\subsection{Where the time goes}
\label{supp:eff_cost}
\begin{table}[t]
\centering\small\setlength{\tabcolsep}{3pt}
\caption{\textbf{Where the 176.2\,s go.} Stage times of the $4096^3$ run in Table~2 (seconds, peak allocated GPU memory of the stage in GiB; concurrent processes are summed).}
\label{tab:supp_cost}
\begin{tabular}{@{}p{0.60\linewidth}rr@{}}
\toprule
Stage & Time (s) & Peak (GiB)\\
\midrule
\raggedright Coarse $512^3$ shape: Pixal3D start and TRELLIS.2 reference frame with their cameras (two concurrent processes) & 76.5 & 23.5\\
\raggedright Start placement (encoding and decoding at $512^3$) & 1.9 & 9.1\\
\raggedright Camera estimation of the start (silhouette fit, refinement, crop) & 35.8 & 10.8\\
\raggedright Detail injection ($512^3$--$4096^3$, 4 levels) & 4.4 & 18.0\\
\raggedright Encoding ($1024^3$--$4096^3$) & 19.4 & 36.2\\
\raggedright Refinement network ($2048^3$, $4096^3$) & 19.6 & 26.4\\
\raggedright Decoding ($2048^3$, $4096^3$) & 8.6 & 48.5\\
\raggedright Mesh export & 6.9 & 39.0\\
Stage hand-over and synchronization & 3.1 & --\\
\midrule
Total & 176.2 & 48.5\\
\bottomrule
\end{tabular}
\end{table}
Table~\ref{tab:supp_cost} decomposes the $4096^3$ run of
Table~2. The two refinement stages together take
19.6\,s, 11\% of the total, and the network at $4096^3$
runs in 14.5\,s: recursive refinement adds three octaves of
resolution at the cost of two network evaluations. Everything that grows
with the target resolution, from detail injection to mesh export, takes
59.0\,s; the coarse $512^3$ shape and its cameras, which do not depend on
the target resolution, take 65\% of the time. The Pixal3D start and the
TRELLIS.2 reference frame used to place it (Sec.~\ref{supp:start}) are
generated in two concurrent processes. Denoising at $4096^3$ needs 26.4\,GiB;
the reported peak of 48.5\,GiB is set by decoding, which keeps the complete
$4096^3$ geometry on the GPU to finish in seconds.

\subsection{Index-based spatial-cell attention}
\label{supp:eff_attn}
\begin{table}[!htb]
\centering\footnotesize\setlength{\tabcolsep}{2.6pt}
\caption{\textbf{Spatial-cell attention on $4096^3$ supports} (one HR attention layer, 12 heads, BF16; median latency in ms, extra peak memory in GiB). All local variants compute the same attention on the same inputs; Ours is bit-identical to Gather\,+\,FA2 (Gather). Global FA2 attends over all HR tokens and is shown for reference.}
\label{tab:supp_attn}
\begin{tabular}{@{}lr|cc|cc|c@{}}
\toprule
 & & \multicolumn{2}{c|}{Latency (ms) $\downarrow$} & \multicolumn{2}{c|}{Memory (GiB) $\downarrow$} & Global\\
Support & Tokens & Ours & Gather & Ours & Gather & FA2 (ms)\\
\midrule
N4 & 143{,}440 & \textbf{4.31} & 5.67 & \textbf{0.82} & 1.92 & 476\\
N6 & 230{,}786 & \textbf{7.21} & 9.27 & \textbf{1.32} & 3.10 & 1239\\
T26 & 307{,}193 & \textbf{11.15} & 13.29 & \textbf{1.76} & 3.99 & 2200\\
P2 & 403{,}109 & \textbf{13.72} & 16.72 & \textbf{2.31} & 5.31 & 3802\\
N8 & 441{,}897 & \textbf{15.41} & 18.45 & \textbf{2.53} & 5.73 & 4587\\
K20 & 456{,}422 & \textbf{14.93} & 18.42 & \textbf{2.61} & 6.06 & 4890\\
SC02 & 602{,}807 & \textbf{18.48} & 23.28 & \textbf{3.45} & 8.04 & 8779\\
A02 & 708{,}592 & \textbf{23.34} & 28.17 & \textbf{4.05} & 9.05 & 12257\\
\bottomrule
\end{tabular}
\end{table}
Table~\ref{tab:supp_attn} isolates the HR local attention
(Sec.~\ref{supp:routing}) on eight $4096^3$ supports with
143k--708k HR tokens. Queries, keys, values, and cell
neighbor sets are captured from the first transformer block, and every
variant runs on exactly these tensors.
\emph{Gather\,+\,FA2} copies each cell's neighbor rows into contiguous
buffers and calls the variable-length FlashAttention-2
kernel~\cite{dao2024flashattention2} once, the standard way to run attention
over irregular neighborhoods.
Our kernel is faster on every support: 1.19--1.31$\times$ faster
than Gather\,+\,FA2, whose output it reproduces bit for bit. It also
needs 55--57\% less memory than Gather\,+\,FA2, because the
1.1--5.0\,GiB of copied keys and values never exist; the copy alone
accounts for 23--27\% of the Gather\,+\,FA2 latency. The HR-to-LR read
uses the same kernel and is 1.13--1.17$\times$ faster than its gather-based
counterpart, again bit-identical. Global FlashAttention-2 over the same
tokens takes 110--525$\times$ longer than our local attention
(Fig.~1, right), so the
spatial-cell decomposition is what makes denoising at $4096^3$ practical.
Index-only reads together with host-side LR caching lower the peak GPU memory
of a $4096^3$ sampling run with 118k HR tokens from 24.0 to 11.8\,GiB, with
bit-identical samples.
\subsection{Fused kernels}
\label{supp:eff_block}
\begin{table}[!htb]
\centering\small\setlength{\tabcolsep}{2.6pt}
\caption{\textbf{Fused kernels}, one transformer block (median ms). PyTorch: original operators; Fused: kernels of Sec.~\ref{supp:kernels}. Element-wise: profiler time of element-wise and normalization kernels.}
\label{tab:supp_block}
\begin{tabular*}{\linewidth}{@{\extracolsep{\fill}}lrccc@{}}
\toprule
Support & Tokens & PyTorch & Fused & Element-wise (ms)\\
\midrule
N4 & 143{,}440 & 123.2 & \textbf{84.8} & 55.1 $\to$ 20.9\\
T26 & 307{,}193 & 276.1 & \textbf{190.6} & 118.7 $\to$ 45.5\\
A02 & 708{,}592 & 647.2 & \textbf{454.3} & 274.3 $\to$ 105.5\\
\bottomrule
\end{tabular*}
\end{table}
Table~\ref{tab:supp_block} measures one complete transformer block,
including HR--LR attention, modulation, image cross-attention, and the
feed-forward network. The kernels of Sec.~\ref{supp:kernels} make the block
1.42--1.45$\times$ faster than the PyTorch implementation. The
gain comes from the memory-bound operators: the profiler time of
element-wise and normalization kernels falls by 2.6$\times$, while
matrix multiplications and attention, which are compute-bound, are left
to their libraries. The fused block matches the unfused one to 0.4--0.5\%
relative $\ell_2$ (0.25\% on the final $4096^3$ latent), mostly from rounding
the merged LoRA weights.

\end{document}